%% file: iclr2027_conference.tex
\documentclass{article} % For LaTeX2e
\usepackage{iclr2027_conference,times}

\input{math_commands.tex}

\usepackage{hyperref}

\definecolor{uclablue}{rgb}{0.15, 0.45, 0.68}
\hypersetup{
    breaklinks,
    citecolor=uclablue,
    colorlinks=true,
}

\usepackage{url}
\usepackage{xspace}
\usepackage{amssymb}
\usepackage{graphicx}
\usepackage{wrapfig}
\usepackage{booktabs}
\usepackage{multirow}
\usepackage{subcaption}
\usepackage{tabularx}

\newcommand{\trace}[0]{\texttt{smartcomment}\xspace} 

\newcommand{\bench}[0]{\texttt{MemTraceBench}\xspace} 
 
\newcommand{\memtrace}{\texttt{MemTrace}\xspace}
\newcommand{\memtracels}{\texttt{MemTrace-OBS}\xspace}

\newcommand{\redbold}[1]{\textcolor{red}{\textbf{#1}}}

\title{MemTrace: Tracing and Attributing Errors \\ in Large Language Model Memory Systems}

\author{
Xinle Deng\textsuperscript{1,2}\footnotemark[1],
~Ruobin Zhong\textsuperscript{1}\footnotemark[1], 
~Hujin Peng\textsuperscript{1}\thanks{Core Contributor.}, 
~Xiaoben Lu\textsuperscript{1},
~Yanzhe Wu\textsuperscript{1},
~Guang Li\textsuperscript{1}, \\
~\textbf{Buqiang Xu}\textsuperscript{1},
~\textbf{Yunzhi Yao}\textsuperscript{1}, 
~\textbf{Jizhan Fang}\textsuperscript{1,2}, 
~\textbf{Haoliang Cao}\textsuperscript{2}, 
~\textbf{Junjie Guo}\textsuperscript{2}, 
~\textbf{Yuanqiang Yu}\textsuperscript{2}, \\
~\textbf{Yuan Yuan}\textsuperscript{1},
~\textbf{Ziqing Ma}\textsuperscript{2}, 
~\textbf{Rui Hu}\textsuperscript{2}, 
~\textbf{Baohua Dong}\textsuperscript{2},
~\textbf{Hangcheng Zhu}\textsuperscript{2},
~\textbf{Ningyu Zhang}\textsuperscript{1}\thanks{Corresponding Author.}\\
\textsuperscript{1}Zhejiang University,
~\textsuperscript{2}Alibaba Group\\
\texttt{\{dengxinle, zhangningyu\}@zju.edu.cn} \\
}

\iclrfinalcopy % Uncomment for camera-ready version, but NOT for submission.
\begin{document}

\maketitle

\begin{abstract}
Memory is essential for enabling large language models to support long-horizon reasoning, yet existing memory systems remain unreliable and difficult to debug. Tracing memory's dynamic evolution is crucial to understand how information is synthesized, propagated, or corrupted over time. In this work, we study the new problem of error tracing and attribution in LLM memory systems. We propose a novel framework that transforms memory pipelines into executable memory evolution graphs, enabling fine-grained tracing of operational information flow. We then construct \bench, a benchmark collected from representative memory systems such as Long-Context, RAG, Mem0, and EverMemOS, to systematically study memory failure modes. We further introduce an automatic attribution method that iteratively traces operation subgraphs to pinpoint the root cause of any failed case. Our analysis reveals that memory failures are systematic, stemming from operation-level issues like information loss and retrieval misalignment. Crucially, we leverage these fine-grained attribution signals to guide downstream prompt optimization, establishing a closed-loop system that automatically corrects faults and boosts end-task performance by up to 7.62 percentage points\footnote{\url{https://github.com/zjunlp/MemTrace}.}.
\end{abstract}

%% Please note that we have introduced automatic line number generation
%% into the style file for \LaTeXe. This is to help reviewers
%% refer to specific lines of the paper when they make their comments. Please do
%% NOT refer to these line numbers in your paper as they will be removed from the
%% style file for the final version of accepted papers.

\input{section/1.intro}

\input{section/2.preliminary}
\input{section/3.dataset}
\input{section/4.method}

\input{section/5.experiments}
\input{section/6.application}
\input{section/7.related_work}
\input{section/8.conclusion}

\subsection*{AI use statement}
Generative AI is used both as an experimental component and as a limited research and writing assistant. As detailed in the paper, LLMs are used throughout the experimental pipeline for memory system execution, failure attribution, evaluation, prompt optimization, and diagnostic report generation. MemTraceBench builds on public datasets containing LLM-synthesized interaction trajectories. During benchmark construction, three human annotators independently provide labels and explanations for each failure case. After the final error type and faulty operation have been determined by the annotators, an LLM is only used to aggregate and polish their explanations. Generative AI tools are additionally used for minor code completion, editing, debugging, and language polishing of the manuscript. They are not used to formulate the research question, design the methodology, conduct the literature review, or determine the final conclusions. All AI-assisted text, code, annotations, and other artifacts included in this work are manually reviewed. We take responsibility for the final content of this work, including text, claims or artifacts produced with the aid of generative AI.

\subsection*{Ethics statement}
\memtrace is intended to support failure diagnosis and transparency for non-parametric memory systems. In this work, \bench is constructed from publicly available, fully LLM-synthesized user trajectories and does not contain real user interactions. However, applying \memtrace to real-world memory systems may involve execution traces that contain sensitive user information \citep{Chen2026MemPrivacyPP}. Such use requires careful data governance, including informed consent, access control, anonymization when possible, and secure storage of logs and generated reports. It is also critical to acknowledge that the diagnoses produced by \memtrace may be imperfect. Consequently, they should be treated as assistive evidence rather than definitive judgments, with human review required before drawing reliability conclusions or deploying system changes.

\subsection*{Reproducibility statement}
All datasets and models used in this work are publicly accessible through their respective official sources or APIs. Details of the data construction process are provided in Appendix~\ref{app:dataset_construction}, and Appendix~\ref{app:smartcomment_example} presents a concrete example of inserting \trace statements into the source code of a memory system. The main text provides the key implementation details of our experiments. Due to space constraints, additional implementation details, including experimental configurations and procedures, are provided in Appendices~\ref{app:failure_attribution_details}, \ref{app:report_generation}, and \ref{app:opt-details}. Upon acceptance, we will release the dataset, source code, annotation website, and prompts used in this work on GitHub. These resources will enable researchers to reproduce our experimental results and facilitate further research on automatically identifying and diagnosing errors in memory systems.

\bibliography{iclr2027_conference}
\bibliographystyle{iclr2027_conference}

\appendix
\input{section/appendix}

\end{document}

%% file: math_commands.tex
\usepackage{amsmath,amsfonts,bm}

\def\eqref#1{equation~\ref{#1}}
\def\1{\bm{1}}

\DeclareMathAlphabet{\mathsfit}{\encodingdefault}{\sfdefault}{m}{sl}
\SetMathAlphabet{\mathsfit}{bold}{\encodingdefault}{\sfdefault}{bx}{n}

%% file: section/1.intro.tex
\section{Introduction}

Memory systems are a core component of large language model (LLM) agents, enabling them to evolve from isolated task solvers into stateful systems capable of long-horizon tasks and continual learning \citep{Xu2025AMEMAM, Fang2025LightMemLA, Yang2026PlugMemAT, Cao2025RememberMR, Wang2025MIRIXMM}. 
By retaining information across interactions, updating state over time, and leveraging past experience for future decisions, memory has become widely adopted in applications such as personalized assistants and coding agents \citep{DBLP:conf/uist/ParkOCMLB23, DBLP:conf/nips/LiXSCJN24, DBLP:conf/cvpr/YangLGDZZWZXWOL25, DBLP:conf/emnlp/XiongCKZS25, DBLP:conf/iclr/0001LSXTZPSLSTL25, Wang2025UITARS2TR, Wen2025AIFS}.
However, as memory systems become increasingly complex, a fundamental question remains underexplored: \textbf{when a memory-augmented agent fails, how can we identify where the error originates?}

% \begin{figure}[t]
%     \centering
%     \includegraphics[width=\columnwidth]{figures/overview.pdf}
%     \caption{\textbf{Framework for automatic diagnosis of LLM memory systems.}
% We first execute a memory system to construct an execution graph. 
% Given a failed case, \memtrace performs step-by-step tracing over this graph to locate the faulty operation. 
% This framework is general across different memory systems and enables faster failure attribution than human experts.}
%     \label{fig:intro_overview}
% \end{figure}

Compared to prior work on diagnosing stateless agentic systems 
\citep{Baker2025MonitoringRM, DBLP:conf/icml/ZhangY0LHZL0W0W25, Wang2026FromFL, li2026codetracer}, 
failure attribution in LLM memory systems presents a distinct challenge.
In stateless agents, failures are often localized within the current execution trajectory, such as an incorrect tool call, retrieval result, or reasoning step.
In contrast, memory-augmented agents maintain persistent states across interactions, so failures may originate from earlier memory construction, update, or deletion operations and only surface much later during retrieval or response generation.
For example, a user preference may be correctly stored at first but later overwritten by an incorrect update, causing a downstream failure far removed from its origin.
Such failures are difficult to diagnose from chronological traces alone.
A linear execution log records operations in order, but as a flat sequence from different parts of the memory pipeline, it lacks the structure \citep{DBLP:conf/chi/JiangSX23} needed to show how memory variables are created, modified, overwritten, propagated, and finally used in a failed prediction.
Existing memory benchmarks
\citep{DBLP:conf/acl/MaharanaLTBBF24, DBLP:conf/iclr/WuWYZCY25, Chen2025HaluMemEH, Bian2026RealMemBL}
are similarly outcome-oriented: they can reveal whether a system successfully
stores, retrieves, or uses relevant information, but they are not designed to
recover the causal path by which a failure is introduced and propagated.
This exposes a key traceability gap in LLM memory systems: failures are observable, yet the faulty operations, their introduction time, and their propagation paths remain difficult to identify. \looseness=-1

\begin{wrapfigure}{r}{0.51\columnwidth}
    \vspace{-0.0cm}
    \includegraphics[width=1.0\linewidth]{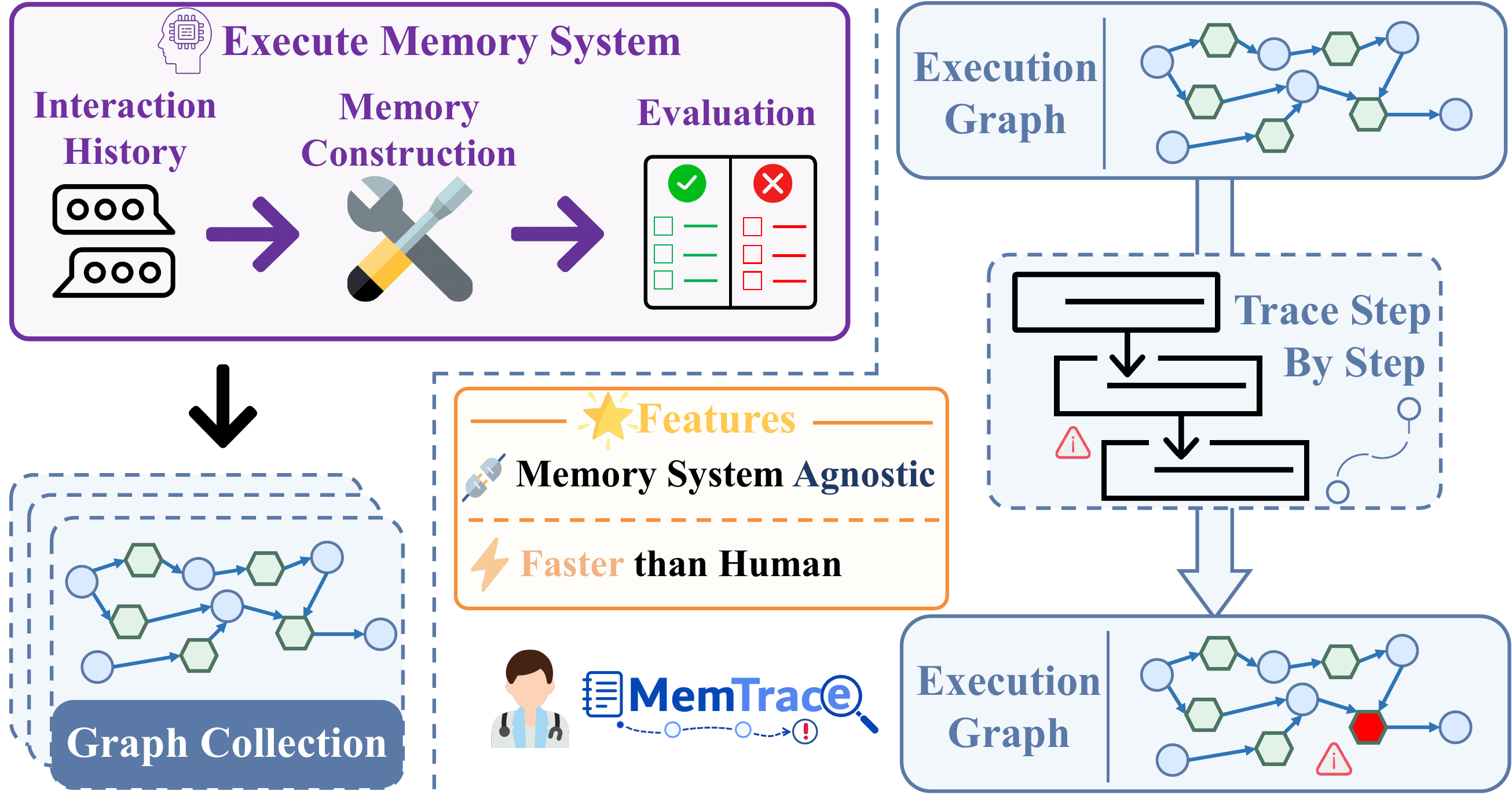}
    \vspace{-0.5cm}
    \caption{\textbf{Framework for automatic diagnosis of LLM memory systems.}
    We first execute a memory system to construct an execution graph. 
    Given a failed case, \memtrace performs step-by-step tracing over this graph to locate the faulty operation. 
    This framework is general across different memory systems and enables faster failure attribution than human experts.}
    \label{fig:intro_overview}
    \vspace{-0.2cm}
\end{wrapfigure}

To address this problem, we propose a novel framework for error tracing and attribution in LLM memory systems, as shown in Figure~\ref{fig:intro_overview}.
\textbf{Our key idea is to expose memory-system execution as a unified operation-variable graph through a system-agnostic tracing toolkit.}
This execution graph records memory operations and their associated variables, and connects variables through shared operations to reveal information flow during memory construction, update, retrieval, and reasoning.
Unlike chronological logs, the graph explicitly captures which operations consume, modify, overwrite, or propagate each memory variable, allowing attribution to follow information dependencies across turns and sessions.
Based on this representation, we introduce three contributions. 
First, we define a structured error taxonomy grounded in execution graph patterns. 
Second, we construct \bench, a diagnostic benchmark with human-annotated 160 real failure cases from four memory systems and three public datasets, each including question-answer (QA) pairs, execution logs, ground-truth error labels, faulty operations, and human explanations. 
Third, we propose \memtrace, an automatic attribution method that operates directly on execution graphs: given a failure case, it retrieves relevant source messages and then traces information-flow subgraphs to locate the decisive faulty operation.
Extensive experiments on \bench show that diagnosing failures in memory systems remains challenging. 
However, \memtrace can successfully recover meaningful faulty operations and error types, and generate explanations for system debugging. 
Beyond error analysis, its attribution signals can guide automatic prompt optimization, improving end-task performance by up to 7.62 percentage points.

%% file: section/2.preliminary.tex
\section{Tracing and Attributing Errors in Memory Systems}
\label{sec:problem_formulation}

Automatic failure attribution consists of two steps: collecting the system’s execution trace and analyzing it to localize the failure source.
In this work, we use $\mathcal{M}$ to denote a non-parametric memory system that processes a trajectory $\tau$ and answers question $q$ with a prediction $\hat{a}$ and a golden answer $a$. Its execution consists of memory updates $\mathcal{U}_{\mathcal{M}}$, memory reads $\mathcal{R}_{\mathcal{M}}$, and answer generation $\mathcal{Q}$. See Appendix~\ref{app:memory_system_definition} for the full formalization.

\paragraph{Background.} Concretely, we instrument the source code of the memory system and execute it on a trajectory $\tau$ and question $q$. Whenever the system performs a memory update $\mathcal{U}_{\mathcal{M}}$, a memory read $\mathcal{R}_{\mathcal{M}}$, or an answer generation step $\mathcal{Q}$, we use a toolkit (Details in Appendix \ref{app:Tracing Toolkit}) to automatically record the involved variables (e.g., the input question $q$ and the predicted answer $\hat{a}$), the operations applied to them, and the dependency relations among them. 
This process produces an execution graph $\mathcal{G} = (\mathcal{V}, \mathcal{O}, \mathcal{E})$, where $\mathcal{G}$ is a directed acyclic bipartite graph. The node set consists of variables $\mathcal{V}$ and operations $\mathcal{O}$. Variables represent concrete artifacts produced during execution, such as raw messages, retrieved memory units, intermediate summaries, and prompts. 
Operations represent computation steps, such as LLM inference, tool invocation, retrieval, filtering, or parsing functions. 
The directed edges $\mathcal{E}$ capture information flow between variables and operations. Each operation $o \in \mathcal{O}$ takes a subset of variables as inputs, denoted as $\mathrm{In}(o) \subset \mathcal{V}$, and produces a subset of variables as outputs, denoted as $\mathrm{Out}(o) \subset \mathcal{V}$. Finally, we define a binary outcome indicator $Z(\mathcal{G}) \in \{0, 1\}$, where $Z(\mathcal{G}) = 1$ indicates that system fails to answer the question, and $Z(\mathcal{G}) = 0$ indicates success. 
In practice, this outcome can be obtained by comparing the prediction $\hat{a}$ with the golden answer $a$ based on an LLM.

\paragraph{Problem Definition.} Given a failed execution graph $\mathcal{G}$ for question $q$ together with the golden answer $a$, our objective is to identify the \textbf{earliest decisive faulty operation}, denoted as $o^*$. 
For an operation $o$, let $\mathrm{Anc}_{\mathcal{G}}(o)$ and $\mathrm{Desc}_{\mathcal{G}}(o)$ denote its strictly upstream ancestor operations and strictly downstream descendant operations in the execution graph, respectively.
An operation $o$ is a candidate earliest error if it is faulty and all operations in $\mathrm{Anc}_{\mathcal{G}}(o)$ are functionally correct. 
To determine whether it is decisive, we construct an idealized execution\footnote{The idealized execution is used only to specify what counts as a decisive faulty operation in our problem definition. It is not a replay step performed in practice. Replaying all affected downstream operations is prohibitively expensive, especially for long-horizon interactions, and may not faithfully realize the intended counterfactual because downstream modules can still fail even after the upstream error has been corrected.}. 
Specifically, $\mathcal{G}^{(o, *)}$ replaces the faulty outputs of $o$ with their correct counterparts and assumes ideal execution for every operation in $\mathrm{Desc}_{\mathcal{G}}(o)$. 
The operation is decisive if this intervention rescues the failed execution, i.e., $Z(\mathcal{G}^{(o, *)})=0$.
Let $\mathcal{D}(\mathcal{G})$ denote the set of operations satisfying this criterion. 
Using the execution timestamp $t_o$ only to order decisive candidates, we then define\footnote{Real-world failures may involve multiple independent faulty operations. Our primary focus is the singleton setting, which allows us to isolate the core problem of attributing a failure to its earliest decisive faulty operation. This setting is already challenging in practice and provides a natural foundation for more general joint attribution. We discuss possible extensions of \memtrace to the non-singleton setting in Appendix~\ref{app:multi_error_attribution}.}
\begin{equation}
o^* = \operatorname*{arg\,min}_{o \in \mathcal{D}(\mathcal{G})}\; t_o .
\end{equation}
Although our failure attribution objective follows prior work on LLM agent systems \citep{DBLP:conf/icml/ZhangY0LHZL0W0W25, Zhang2025AgenTracerWI, Wang2026FromFL}, the memory-system setting studied here differs in several important ways. 
In prior works, the execution trace is often treated as a relatively short sequence of logs produced by a single task run. In contrast, a memory system is executed over a long historical trajectory $\tau$, so its trace can grow to tens of megabytes in our setting. More importantly, the trace is inherently not an unstructured, flat log. 
For example, a memory unit produced by an earlier memory update may later be retrieved, transformed, or used in answer generation, creating dependencies that span both different operations and different time steps.

% Footnote: In real-world systems, a failure may be jointly caused by multiple independent faulty operations. In such cases, singleton attribution is insufficient. We discuss this non-singleton setting and possible extensions of \memtrace in Appendix~\ref{app:multi_error_attribution}

%% file: section/3.dataset.tex
\begin{figure*}[t]
    \centering
    \includegraphics[width=\textwidth]{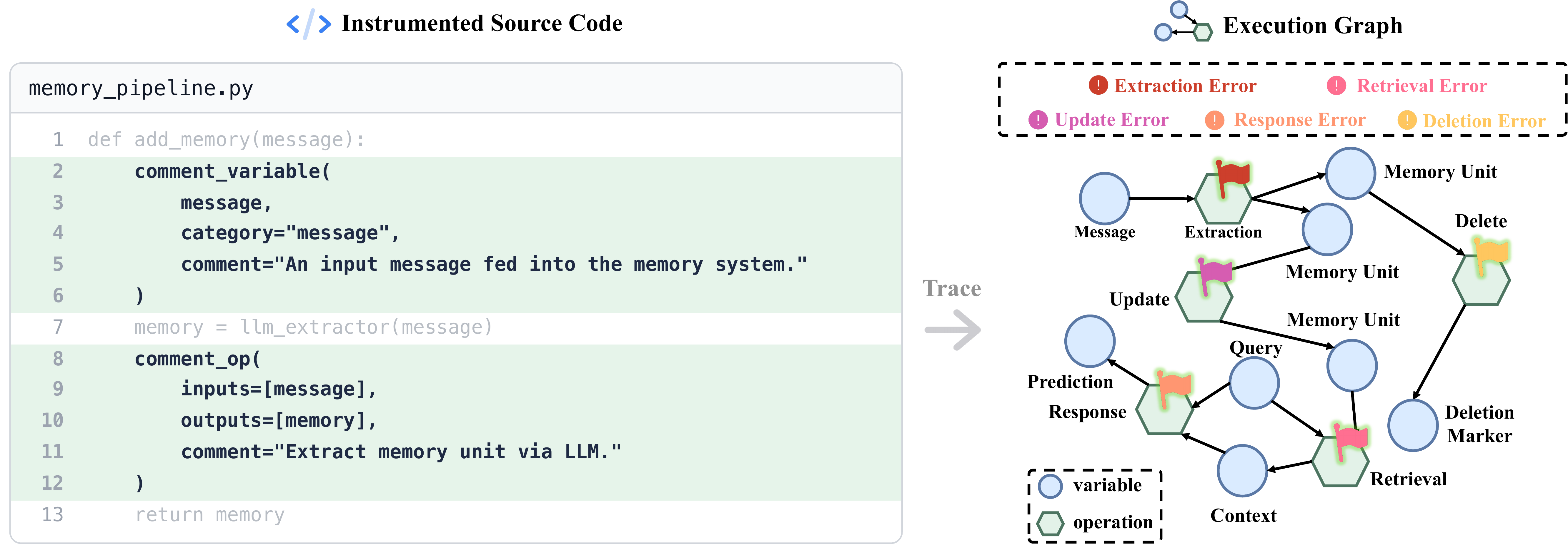}
    \caption{
    \textbf{The overview of dataset construction process.}
    We define seven error types, five of which are specific to memory systems. We insert \trace-related code into each memory system and run these systems on public datasets to collect execution graphs and failed cases. For each failed case, once our annotators confirm that it is not caused by annotation errors or LLM-as-a-Judge errors, they identify the earliest decisive faulty operation in the execution graph and provide its cause and error type.
    }
    \label{fig:annotation-overview}
\end{figure*}

\section{\bench Construction}

Due to the lack of datasets for evaluating automatic failure attribution in stateful agents with non-parametric memory, we construct a new dataset, \bench (MIT Licence). 
Figure~\ref{fig:annotation-overview} in Appendix illustrates the overview of construction process. Each example in \bench includes a question, its corresponding golden answer, the full execution trace of the system, and annotated failure information. 
The annotations include the unique identifiers of faulty operations, their error types, and explanations. 
We construct our benchmark using question–answer pairs from LoCoMo \citep{DBLP:conf/acl/MaharanaLTBBF24}, LongMemEval \citep{DBLP:conf/iclr/WuWYZCY25}, and RealMem \citep{Bian2026RealMemBL}. 
Four representative memory systems are selected, including long-context memory, RAG \citep{DBLP:conf/nips/LewisPPPKGKLYR020}, Mem0 \citep{Chhikara2025Mem0BP}, and EverMemOS \citep{Hu2026EverMemOSAS}. See Appendix~\ref{app:data_sources} for further details of data sources and memory systems.

Constructing this benchmark requires collecting fine-grained execution graphs for memory systems, rather than only the inputs and outputs of LLM calls. 
In particular, we need to capture how messages produce memory units, how memories evolve over time, and how intermediate variables depend on one another across memory construction, retrieval, response generation, and evaluation. Since existing memory systems use heterogeneous schemas and code structures, we collect traces through explicit instrumentation rather than rewriting them around a unified abstraction. 
Moreover, existing instrumentation-based tracing frameworks are mostly event-centric and do not directly track variable evolution and dependencies. 
We therefore develop \trace, a lightweight tracing package for recording developer-specified operations, variables, and their dependencies. 
We instrument each memory system by adding tracing statements at key operations and then run the instrumented systems on sampled trajectories, collecting 1,514 distinct errors across all systems (see Appendix~\ref{app:collection_details} for more details). 
\textbf{We then recruit five annotators from the author team to identify the faulty operations, and provide corresponding error types and explanations.}
The final benchmark contains 160 system-related failure cases. Appendix~\ref{app:annotation_process} shows the annotation process. 
We also provide more details on \trace in Appendix \ref{app:Tracing Toolkit}. 

%% file: section/4.method.tex
\begin{figure*}[t!]
\centering
\resizebox{0.9\textwidth}{!}{
\includegraphics{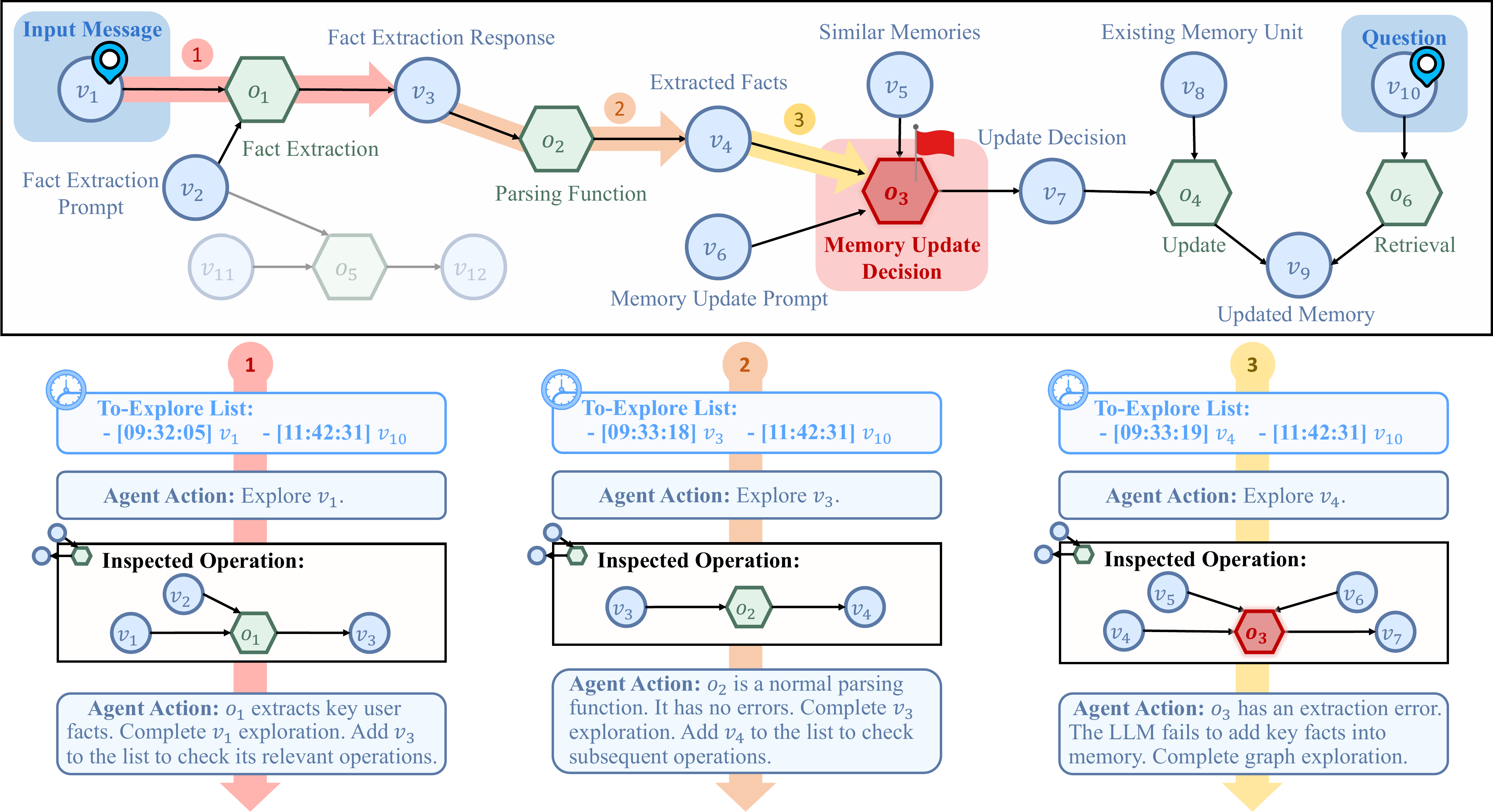}}
\caption{
\textbf{An illustrative workflow of \memtrace.}
The initial to-explore list contains $v_1$ and $v_{10}$. Starting from $v_1$, the agent inspects the operation subgraph corresponding to the operation $o_1$, and finds that $o_1$ correctly extracts the key user facts. The agent then adds $v_3$ to the list to inspect subsequent operations. By continuing this graph exploration process, the agent identifies the faulty operation $o_3$ in the third iteration. 
}
\label{fig:method}
\end{figure*}

\section{Methodology}
We propose \memtrace to attribute failures in non-parametric memory systems. It casts failure attribution as an agentic graph exploration problem. As illustrated in Figure~\ref{fig:method}, the agent iteratively inspects local operation subgraphs in $\mathcal{G}$ and updates its exploration state until it identifies the target decisive error $o^*$ or reaches the maximum number of reasoning steps. At each iteration, \memtrace maintains a bounded to-explore list of size at most $N$. The list is implemented as a priority queue over variable nodes. Each variable $v \in \mathcal{V}$ is associated with its insertion timestamp $t_v$ in the execution graph. Variables with earlier timestamps are assigned higher priority in the list. This priority ensures the agent inspects earlier operations first. 
The overall method contains three modules: selecting starting points, exploring the execution graph, and managing the agent's working context.

\subsection{Initialization of Starting Point}

Before exploring the graph, \memtrace needs to choose a small set of starting variables. A naive strategy is to initialize the to-explore list with all system inputs, including the question $q$ and all raw input messages $\{m_i\}_{i = 1}^{n}$ in the historical trajectory $\tau$. However, this creates a very large search space, especially when the trajectory spans many sessions. 

To reduce the initial branching factor, \memtrace uses hybrid retrieval to identify source messages that are most likely to contain the critical information needed by the failed question. Specifically, we construct a retrieval query by concatenating the question with the golden answer. We then perform both dense retrieval and sparse retrieval over the raw message set $\{m_i\}_{i=1}^{n}$ to obtain top-$N$ candidate messages from each retriever. The two ranked lists are fused by Reciprocal Rank Fusion (RRF), and the top $\lfloor N/2 \rfloor$ messages from the fused ranking are selected. Finally, these messages together with the question $q$ are used to create the initial to-explore list $\mathcal{L}_0$. Note that we reserve the remaining capacity so that the agent can add newly discovered downstream variables during exploration.

\subsection{Execution Graph Exploration}

At the $t$-th iteration, given the current to-explore list $\mathcal{L}_{t-1}$, \memtrace selects the variable $v_t$ with the earliest timestamp and marks it as the variable under exploration. It then retrieves all operations directly involving this variable:
\begin{equation}
    \mathcal{O}(v_t) = \{o \in \mathcal{O} \mid v_t \in \mathrm{In}(o) \cup \mathrm{Out}(o)\}.
\end{equation}
For each operation $o \in \mathcal{O}(v_t)$, \memtrace converts the corresponding operation-level subgraph $\mathcal{G}_o$ into a textual representation, including the operation name, category, comment, input variables, output variables and dependency relations. This localized view allows the agent to reason over the part of the execution graph that is immediately relevant to the current variable, instead of loading the entire graph into context. The agent judges each inspected operation according to the decisive-error criterion defined in Section~\ref{sec:problem_formulation}.  If an operation is locally correct, the agent follows the information flow downstream by adding relevant downstream variables of the operation subgraph into the list:
\begin{equation}
    \mathcal{L}_{t} = (\mathcal{L}_{t - 1} \setminus \{v_t\})  \cup  \mathcal{A}_t,
\end{equation}
where $\mathcal{A}_t \subseteq \mathcal{V}$ is the set of newly selected variables to explore next. This process encourages \memtrace to track the lifetime of critical information through the memory system. The exploration terminates when the agent identifies $o^{*}$, or when the maximum number of reasoning iterations is reached.

% Although our implementation targets singleton decisive error sets, the same framework can be extended to non-singleton cases. Once the agent identifies a decisive faulty operation, it can record it and prune operations strictly downstream of it, since these operations cannot belong to the same earliest decisive error set. The agent can then continue from the remaining to-explore list to search for additional independent faulty operations in the pruned execution graph. We leave this extension to future work.

\subsection{Working Context Management}

Execution graphs for memory systems can be large, often spanning many operations and long variable values. Therefore, \memtrace explicitly manages the agent's working context during graph exploration. From the action space of the agent, \memtrace supports a lightweight preview mode for each operation subgraph. In this mode, concrete variable values are omitted. The agent can then selectively inspect only the variables that are relevant to its current hypothesis. For large variable values, \memtrace provides targeted access through pagination and regex search. The textual representation of operation subgraphs can also be paginated. These tool-level controls prevent sudden context expansion. In addition, \memtrace automatically applies working-context summarization when the context exceeds a predefined safety threshold $T$.

\subsection{Search-Based Operation Exploration}

The graph-based exploration strategy in \memtrace requires the agent to move between variables by following dependency edges, and at each step the agent can only inspect operations involving the current variable. This design can be inefficient when the execution graph is weakly structured or degenerates into a long chain. To handle such cases, we introduce \memtracels.

\memtracels is based on the observation that operation names, variable values, and comments often already reveal the approximate information flow and functional role of each operation. Concretely, it converts each operation-level subgraph into a textual operation block. In this block, dependency edges and unique variable identifiers are removed, while the input variables, output variables, intermediate variables, and operation attributes such as the operation name and comment are preserved. This compressed representation reduces token usage, especially for operations with many repetitive edges\footnote{For example, when retrieving 100 memory units, the query may be connected to every retrieved memory by edges with nearly identical attributes. Representing these edges adds substantial overhead but little additional information.}. We then sort all operation blocks by timestamp and concatenate them with separators to form a weakly structured operation log. Inspired by search mechanisms used by coding agents to navigate large codebases~\citep{DBLP:conf/nips/YangJWLYNP24}, \memtracels equips the agent with a global operation-search tool. Given a regular expression, the tool returns operation blocks whose textual contents match the query, with a configurable limit on the maximum number of returned blocks.

%% file: section/5.experiments.tex
\begin{table*}[t]
\centering
\small
\resizebox{\textwidth}{!}{
\begin{tabular}{llcccccccccc}
\toprule
\multirow{2}{*}{Backbone} & \multirow{2}{*}{Method}
& \multicolumn{2}{c}{Long-Context}
& \multicolumn{2}{c}{RAG}
& \multicolumn{2}{c}{Mem0}
& \multicolumn{2}{c}{EverMemOS}
& \multicolumn{2}{c}{Overall} \\
\cmidrule(lr){3-4}
\cmidrule(lr){5-6}
\cmidrule(lr){7-8}
\cmidrule(lr){9-10}
\cmidrule(lr){11-12}
& & ETA & OIA
  & ETA & OIA
  & ETA & OIA
  & ETA & OIA
  & ETA & OIA \\
\midrule
\multirow{3}{*}{GPT-4.1 mini}
& All-at-Once & 30.00 & 5.00 & 40.00 & 27.50 & 7.50 & 7.50 & 50.00 & 12.50 & 31.88 & 13.13 \\
& \memtracels & 9.17 & 3.33 & 25.83 & 17.5 &  33.33 & 16.67 & 11.67 & 0.0 & 20.00 & 9.38 \\
& \memtrace  & 20.83 & 4.17 & 41.67 & 26.67 & 35.83 & 23.33 & 47.50 & 2.50 & \textbf{36.46} & \textbf{14.17} \\
\midrule
\multirow{3}{*}{GPT-5.4}
& All-at-Once & 40.00 & 0.00 & 80.00 & 52.50 & 35.00 & 5.00 & 52.50 & 17.50 & 51.88 & 18.75 \\
& \memtracels   & 7.50 & 7.50 & 87.50 & 87.50 & 60.00 & 55.00 & 60.00 & 35.00 & 53.75 & \textbf{46.25} \\
& \memtrace  & 20.00 & 20.00 & 72.50 & 65.83 & 70.00 & 59.17 & 55.00 & 7.50 & \textbf{54.38} & 38.13 \\
\bottomrule
\end{tabular}
}
\caption{\textbf{Failure attribution accuracy on \bench across memory systems.} ``ETA'' and ``OIA'' denote the accuracy of error type prediction and faulty operation identification, respectively. All values are reported as percentages. ``Overall'' aggregates results across all memory systems.}
\label{tab:attribution_accuracy}
\end{table*}

\begin{table*}[t]
\centering
\small
\resizebox{\textwidth}{!}{
\begin{tabular}{llcccccccccc}
\toprule
\multirow{2}{*}{Backbone} & \multirow{2}{*}{Method}
& \multicolumn{2}{c}{Long-Context}
& \multicolumn{2}{c}{RAG}
& \multicolumn{2}{c}{Mem0}
& \multicolumn{2}{c}{EverMemOS}
& \multicolumn{2}{c}{Overall} \\
\cmidrule(lr){3-4}
\cmidrule(lr){5-6}
\cmidrule(lr){7-8}
\cmidrule(lr){9-10}
\cmidrule(lr){11-12}
& & Tokens & Time
  & Tokens & Time
  & Tokens & Time
  & Tokens & Time
  & Tokens & Time \\
\midrule
\multirow{3}{*}{GPT-4.1 mini}
& All-at-Once & 1,164.79 & 0.85 & 1,199.22 & 0.85 & 1,204.58 & 0.77 & 1,105.65 & 0.77 & 1,168.56 & \textbf{0.81} \\
& \memtracels   & 692.79 & 2.62 & 684.95 & 1.90 & 1,077.10 & 2.11 & 981.82& 3.63 & \textbf{859.17} & 2.57 \\
& \memtrace  & 4,471.10 & 7.06 & 839.48 & 3.84 & 830.85 & 4.56 & 1,126.10 & 3.82 & 1,816.88 & 4.82 \\
\midrule
\multirow{3}{*}{GPT-5.4}
& All-at-Once & 1,150.14 & 0.83 & 1,199.35 & 0.85 & 1,204.74 & 1.02 & 1,120.69 & 0.87 & 1,168.73 & \textbf{0.89} \\
& \memtracels   & 373.89 & 1.11 & 277.32 & 0.93 & 210.00 & 0.75 & 333.67 & 1.07 & \textbf{298.72} & 0.97 \\
& \memtrace  & 2,524.81 & 5.11 & 1,477.03 & 3.41 & 846.09 & 2.19 & 2,654.01 & 5.63 & 1,875.49 & 4.09 \\
\bottomrule
\end{tabular}
}
\caption{\textbf{Average inference cost on \bench across memory systems.} ``Tokens'' denotes the average token cost (in thousands) required to run automatic failure attribution for one error case, including both total input and output tokens. ``Time'' denotes the average end-to-end runtime (in minutes) for one error case.}
\label{tab:inference_cost}
\end{table*}
\section{Experiments}

\subsection{Experimental Setup}

\paragraph{Baselines and Backbones.} We study failure attribution for stateful agents with non-parametric memory systems, the problem for which no prior baseline is specifically designed (see related discussion in Appendix~\ref{app:prior_failure_attribution_discussion}). We therefore consider All-at-Once \citep{DBLP:conf/icml/ZhangY0LHZL0W0W25}, a general failure-attribution approach that directly analyzes the flattened log. GPT-4.1 mini \citep{GPT41} and GPT-5.4 \citep{GPT54} are used as the agent backbones. More details are provided in Appendix~\ref{app:failure_attribution_details}.

\paragraph{Evaluation Metrics.} We evaluate failure attribution quality using two metrics. \textbf{Error type prediction accuracy} measures whether the agent-predicted error type matches the annotated error type in \bench. \textbf{Faulty operation identification accuracy} measures whether the operation identifier predicted by the agent matches the annotated faulty operation identifier. In addition to attribution accuracy, we report the average token cost and average end-to-end runtime, since practical deployment of automatic failure attribution must handle large volumes of execution logs.

\subsection{Main Results}
\label{sec:main_results}

\begin{figure*}[t!]
\centering
\begin{subfigure}[t]{0.34\textwidth}
  \centering
  \includegraphics[width=\linewidth]{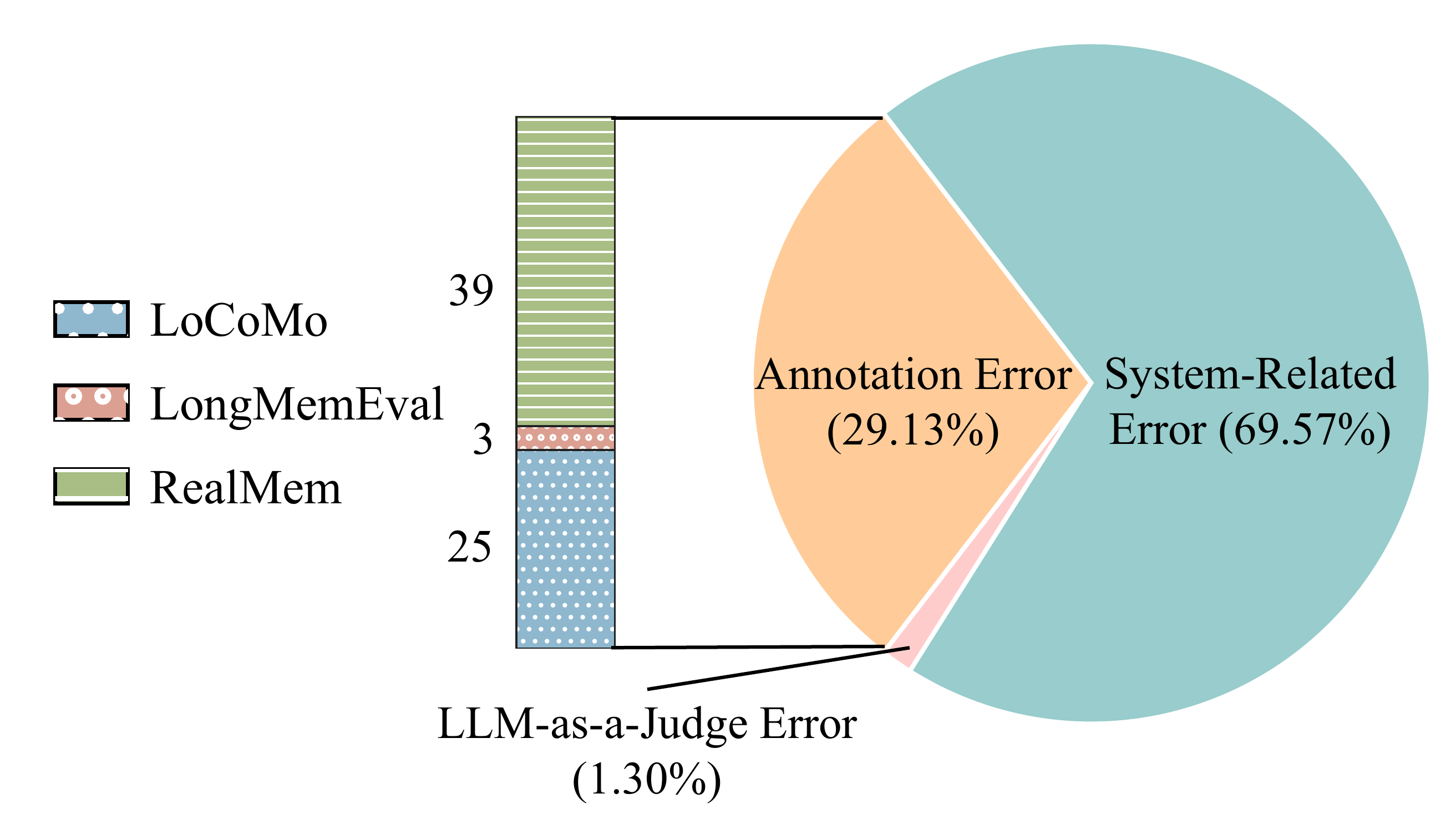}
  \caption{Error distribution across datasets}
  \label{fig:dataset_overview}
\end{subfigure}\hfill
\begin{subfigure}[t]{0.62\textwidth}
  \centering
  \includegraphics[width=\linewidth]{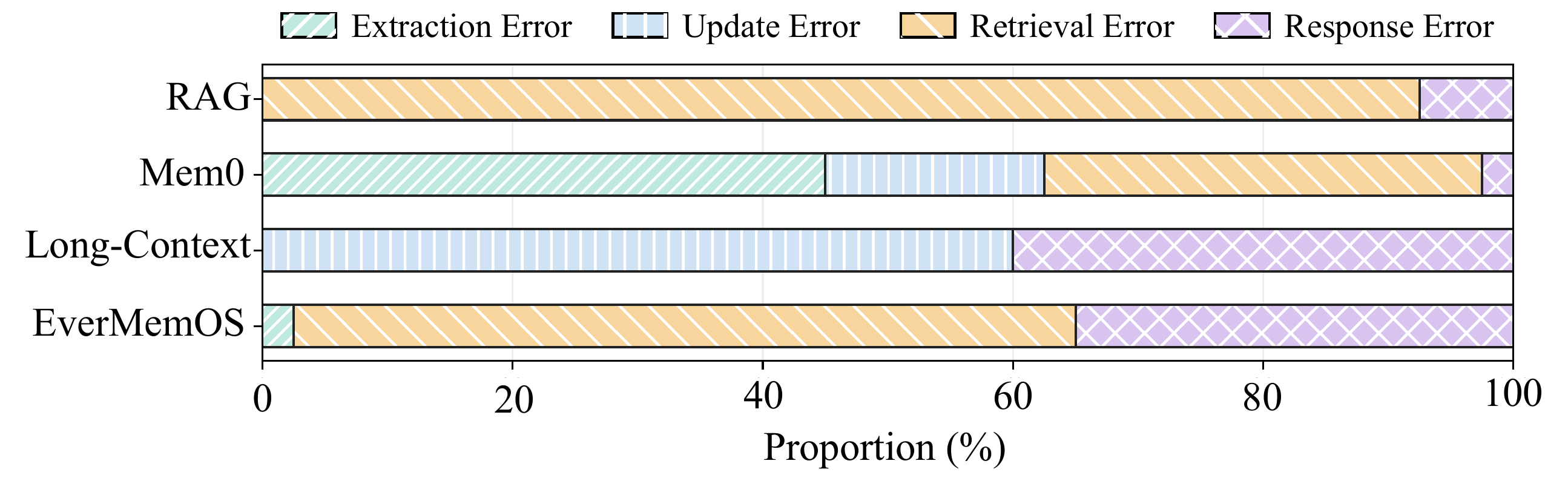}
  \caption{System-related error distribution across memory systems}
  \label{fig:system_error_distribution}
\end{subfigure}

\caption{{\bf Overview of error distribution in \bench.}}
\label{fig:bench_overview}
\end{figure*}

\paragraph{All-at-Once is ill-suited to multi-session stateful memory systems.} All-at-Once achieves only 18.75\% overall OIA on GPT-5.4, substantially below both \memtrace and \memtracels. It also underperforms \memtrace with GPT-4.1 mini. Because it discards many early operations, it favors retrieval and response errors near the end of execution. Its performance therefore varies with the error distributions shown in Figure~\ref{fig:system_error_distribution}.

\paragraph{Graph-based exploration is particularly effective when errors arise during memory construction or when smaller LLMs are used.} As shown in Table~\ref{tab:attribution_accuracy}, \memtrace achieves the highest overall ETA with both backbones, with the clearest improvement under GPT-4.1 mini, where it outperforms \memtracels from 20.00\% to 36.46\%.  We find that search-based operation exploration exhibits a late-stage inspection bias across both backbones. Given global access to operation blocks, the agent often searches for keywords from the golden answer and directly jumps to operations near retrieval or response. The effect of this behavior depends on model capability. GPT-4.1 mini is more susceptible to hard distractors in the retrieved context. It may incorrectly regard retrieval as successful and infer that the missing evidence is lost during extraction. This causes retrieval and response errors to be misclassified as extraction errors. GPT-5.4 more reliably distinguishes decisive evidence from such distractors. Consequently, direct access to late-stage operations is advantageous on RAG and EverMemOS subsets where retrieval and response errors constitute a large proportion of failures. In contrast, failures in Mem0 and Long-Context more frequently originate during memory extraction or update. By explicitly following dependency edges from source messages through memory extraction and subsequent updates, \memtrace preserves information provenance and is better suited to localizing such early-stage faults.

\paragraph{Search-based operation exploration offers an efficient alternative for weakly structured traces.} Table~\ref{tab:inference_cost} shows that \memtracels consistently incurs the lowest overall inference cost across both backbones. It only uses 15.25\% of the tokens and 29.42\% of the runtime required by \memtrace in average on the long-context subset. This advantage is expected because long-context memory performs repeated context updates, producing traces with weak graph structure. By contrast, on the Mem0 subset, \memtracels uses 76.75\% of the tokens and 40.26\% of the runtime of \memtrace on average. RAG and EverMemOS show larger cost reductions than Mem0, likely because both systems maintain and update message buffers before triggering indexing or extraction. This makes parts of their execution graphs locally resemble the structure of long-context memory execution graph. Although \memtrace remains more expensive than \memtracels, it is still substantially faster and cheaper than manual attribution performed by human experts.

\subsection{Analysis}

\paragraph{LLM-identified errors are highly reliable.} Figure~\ref{fig:bench_overview} shows that whenever the LLM judge identifies an error, the verdict is almost always correct. Inspecting the small set of disagreements between the LLM judge and our annotators (see Figure~\ref{fig:judge_error_case}), we find that the judge is overly strict. It penalizes responses that are essentially correct but either overly verbose or lacking sufficient specificity.

\paragraph{High-quality annotation is intrinsically difficult on long-horizon memory benchmarks.} Despite careful human verification, all three datasets contain some annotation errors (see Figures~\ref{fig:annotation_error_case_1} and \ref{fig:annotation_error_case_2}). We find that these errors typically arise from imprecise questions, insufficient source evidence, or inconsistencies between the golden answer and the supporting evidence. In RealMem, the questions and reference answers are substantially more open-ended than those in the other two datasets, which further amplifies annotator subjectivity.

\paragraph{Error distributions reveal distinct bottlenecks across memory systems.}
As shown in Figure~\ref{fig:system_error_distribution}, different memory systems exhibit substantially different failure patterns. 
RAG has no extraction errors because it does not contain an extraction module, whereas Mem0 and EverMemOS both rely on extraction. Notably, EverMemOS produces very few extraction errors. We find that its extraction module is more robust, more generalizable across both human-assistant and multi-party dialogues, and more effective at handling temporal information.
For retrieval errors, Mem0, EverMemOS, and RAG all fail frequently, partly because we retrieve only the top-10 memory units. 
However, this also indicates existing retrieval modules still struggle to recall all target evidence under a limited retrieval budget. 
For EverMemOS, some retrieval failures further originate from the final reranker, which fails to rerank target memories into the top-10 candidates. 
Long-context memory, by design, do not perform retrieval and therefore have no retrieval errors. 
We also observe no deletion errors, likely because deletion is only supported by Mem0 and is rarely tested in current benchmarks\footnote{After analyzing the execution logs of Mem0, we find delete operations account for only 1.02\% of all add, update, and delete operations in Mem0.}. 
Compared with other systems, Mem0 additionally supports memory updates, which leads to more diverse error modes. Finally, all systems exhibit response errors, showing even when related memories are retrieved successfully, effectively using them to give the final answer remains an open challenge.

\begin{wraptable}{r}{0.55\columnwidth}
\centering
\small
\begin{tabular}{lcccc}
\toprule
Method & ETA & OIA & Tokens & Time \\
\midrule
\multicolumn{5}{l}{\textbf{GPT-4.1 mini}} \\
\memtrace & 41.67 & 17.50 & 932.14 & 4.07 \\
+ Source Evidence & 40.55 & 27.22 & 575.69 & \textbf{1.14} \\
+ Prior Knowledge & \textbf{46.39} & 23.89 & 947.32 & 3.69 \\
+ Both & 45.83 & \textbf{29.44} & \textbf{521.80} & 1.43 \\
\midrule
\multicolumn{5}{l}{\textbf{GPT-5.4}} \\
\memtrace & 65.83 & 44.17 & 1,659.04 & 3.74 \\
+ Source Evidence & 69.17 & 54.17 & \textbf{1,036.69} & \textbf{2.41} \\
+ Prior Knowledge & 64.17 & 45.83 & 1,837.41 & 4.94 \\
+ Both & \textbf{70.00} & \textbf{58.33} & 1,475.29 & 3.06 \\
\bottomrule
\end{tabular}
\caption{\textbf{Additional analysis of \memtrace on the subset of \bench.} ``+ Both'' indicates adding both source evidence and prior knowledge.}
\label{tab:memtrace_additional_analysis}
\end{wraptable}

\paragraph{Source evidence and system prior knowledge improve automatic failure attribution.} 
In realistic development settings, practitioners often have access to two forms of auxiliary information: a high-level understanding of the memory-system pipeline, and source evidence provided by the evaluation set for debugging and iteration. We therefore study whether \memtrace can benefit from initializing the to-explore list with source evidence and from adding a coarse pipeline description to the task instruction. This analysis is conducted on a \bench subset that excludes the long-context memory category. As shown in Table~\ref{tab:memtrace_additional_analysis}, source evidence significantly improves OIA as it provides accurate starting points for graph exploration. 
It also reduces attribution cost, since each question-answer pair usually contains only a small number of golden source-evidence messages, typically one to four. 
This keeps the initial to-explore list small. Adding prior knowledge about the memory pipeline also improves OIA.
However, it increases token cost due to additional system-level prompts.
Combining both sources yields the best attribution performance with lower token usage and runtime than the original setting.

\paragraph{Golden answers provide effective signals for retrieving source evidence.}
We further examine whether concatenating the original question with the golden answer can effectively retrieve source evidence for initializing graph exploration. We compare three retrieval strategies: sparse retrieval, dense retrieval, and hybrid retrieval used by \memtrace. We evaluate retrieval performance using Recall@8. As shown in Table~\ref{tab:retrieval_recall}, hybrid retrieval consistently achieves the best performance across all datasets. On LoCoMo and LongMemEval, even the weakest method achieves over 70\%. On RealMem, dense retrieval underperforms sparse retrieval, possibly because its messages are longer, a setting in which sparse retrieval can be more reliable~\citep{DBLP:journals/tacl/LuanETC21}. Overall, these results demonstrate golden answers can provide high-quality starting points for \memtrace exploration. Appendix~\ref{app:retrieval_query_construction} further analyzes source-evidence retrieval when golden answers are unavailable.

% \begin{figure*}[t!]
% \centering
% \begin{minipage}[c]{0.43\textwidth}
%   \centering
%   \begin{subfigure}[t]{\linewidth}
%     \centering
%     \includegraphics[width=\linewidth]{figures/dataset_overview.pdf}
%     \caption{Error Sources}
%     \label{fig:dataset_overview}
%   \end{subfigure}

%   \vspace{0.8em}

%   \begin{subfigure}[t]{\linewidth}
%     \centering
%     \includegraphics[width=\linewidth]{figures/system_error_distribution.pdf}
%     \caption{System-Related Errors}
%     \label{fig:system_error_distribution}
%   \end{subfigure}
% \end{minipage}\hfill
% \begin{minipage}[c]{0.48\textwidth}
%   \centering
%   \begin{subfigure}[t]{\linewidth}
%     \centering
%     \includegraphics[
%       width=\linewidth,
%       height=0.36\textheight,
%       keepaspectratio
%     ]{figures/graph_token_distribution_by_system.pdf}
%     \caption{Token Distribution}
%     \label{fig:graph_token_distribution_by_system}
%   \end{subfigure}
% \end{minipage}
% \caption{{\bf Overview of \bench.}
% (a) Error source distribution and dataset-source breakdowns for annotation errors.
% (b) Error type distribution for each memory system.
% (c) Token distribution of execution graph logs for each memory system.}
% \label{fig:bench_overview}
% \end{figure*}

%% file: section/6.application.tex
\section{Application}

\subsection{Diagnostic Report for Memory Systems}

%Systematically analyzing memory-system failures is often labor-intensive, since execution logs can span long construction, update, retrieval, and response pipelines. 
%MemTrace enables operation-level aggregation of failed cases, allowing us to automatically summarize where and how memory systems fail across the full pipeline. 

Analyzing memory-system failures is labor-intensive due to long construction, update, retrieval, and response pipelines. 
MemTrace enables operation-level aggregation of failures, automatically summarizing where and how systems fail across the pipeline.
We apply this analysis to Mem0 and EverMemOS. 
Tables~\ref{tab:mem0-error-analysis}--\ref{tab:mem0-error-analysis-continued-2} and Tables~\ref{tab:evermemos-error-analysis}--\ref{tab:evermemos-error-analysis-continued-1}
present the generated reports for Mem0 and EverMemOS, respectively. The report generation details are presented in Appendix~\ref{app:report_generation}.
The generated reports reveal different failure patterns in Mem0 and EverMemOS.
For Mem0, the extraction module tends to keep high-level user information while dropping fine-grained details, consistent with \citet{Hu2026CloneMemBL}. 
The report further identifies timestamp reassignment during updates, where content remains unchanged but its time is modified. 
For EverMemOS, no major extraction errors are observed, but aggregation and counting failures appear in the response stage. 
It also localizes issues to specific retrieval components, including the reranker, sufficiency checker, and query reformulation module. 
\textbf{Overall, \memtrace reliably captures overall failure patterns and enables fine-grained diagnosis of concrete pipeline components.}

% \begin{table}[t]
% \centering
% \small
% \begin{tabular}{lcccc}
% \toprule
% Method & ETA & OIA & Tokens & Time \\
% \midrule
% \multicolumn{5}{l}{\textbf{GPT-4.1 mini}} \\
% \memtrace & 41.67 & 17.50 & 932.14 & 4.07 \\
% + Source Evidence & 40.55 & 27.22 & 575.69 & \textbf{1.14} \\
% + Prior Knowledge & \textbf{46.39} & 23.89 & 947.32 & 3.69 \\
% + Both & 45.83 & \textbf{29.44} & \textbf{521.80} & 1.43 \\
% \midrule
% \multicolumn{5}{l}{\textbf{GPT-5.4}} \\
% \memtrace & 65.83 & 44.17 & 1,659.04 & 3.74 \\
% + Source Evidence & 69.17 & 54.17 & \textbf{1,036.69} & \textbf{2.41} \\
% + Prior Knowledge & 64.17 & 45.83 & 1,837.41 & 4.94 \\
% + Both & \textbf{70.00} & \textbf{58.33} & 1,475.29 & 3.06 \\
% \bottomrule
% \end{tabular}
% \caption{\textbf{Additional analysis of \memtrace on the subset of \bench.} ``+ Both'' indicates adding both source evidence and prior knowledge.}
% \label{tab:memtrace_additional_analysis}
% \end{table}
\begin{wrapfigure}{r}{0.60\columnwidth}
\centering
\begin{subfigure}[t]{0.55\linewidth}
  \centering
  \includegraphics[width=\linewidth]{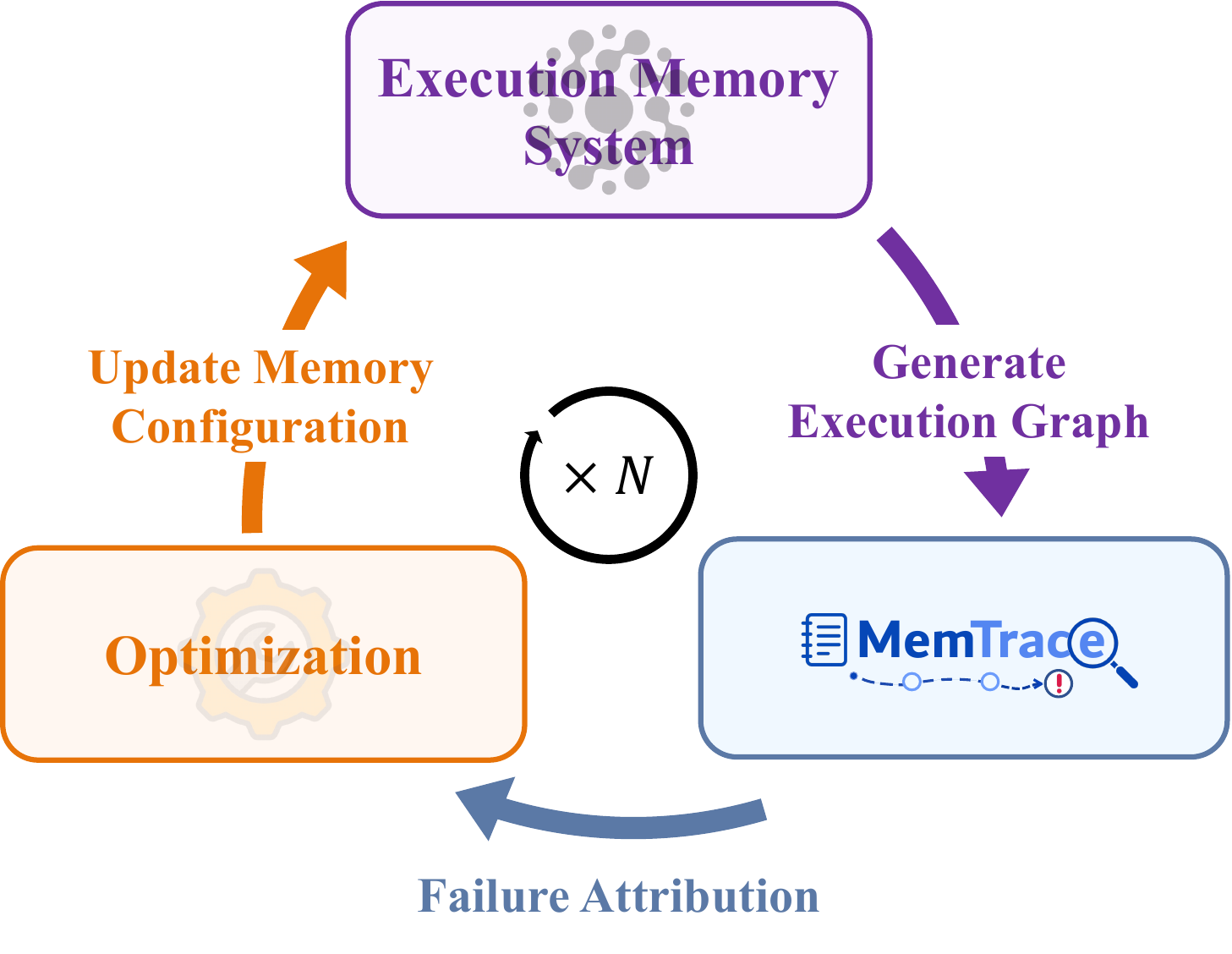}
  \caption{Pipeline}
  \label{fig:optim_overview}
\end{subfigure}\hfill
\begin{subfigure}[t]{0.40\linewidth}
  \centering
  \includegraphics[width=\linewidth]{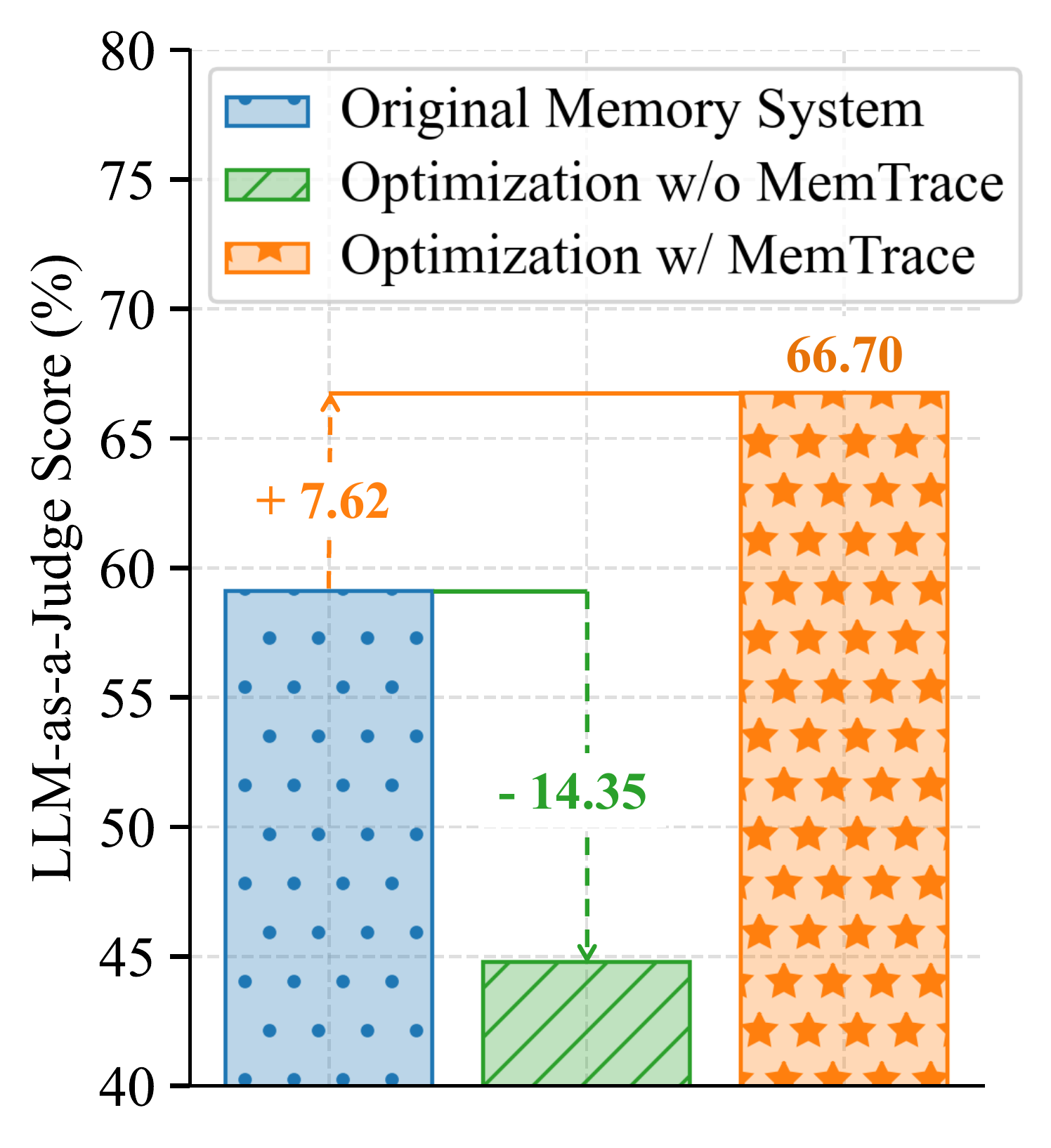}
  \caption{Performance}
  \label{fig:optim_performance}
\end{subfigure}
\caption{{\bf Automatic optimization of Mem0.} 
(a) Overview of the optimization pipeline. (b) Performance comparison of Mem0 before and after optimization, showing an improvement of 7.62 percentage points after three rounds.}
\label{fig:mem0_optimization}
\end{wrapfigure}

\subsection{Automatic Optimization of Prompts in Memory Systems}

Non-parametric memory systems rely on many hand-written prompts, making manual tuning costly. A natural approach is automatic prompt optimization, but existing methods fail in multi-session settings due to long, cross-session execution traces (see Appendix~\ref{app:prompt-opt}). We decouple \emph{credit assignment} from \emph{prompt rewriting} with \trace and \memtrace. 
As illustrated in Figure \ref{fig:optim_overview}, \trace records the runtime execution graph of the memory system, and \memtrace performs credit assignment on this graph to localize the earliest decisive faulty operation. \textbf{Once that operation is identified, prompt optimization reduces to a local problem: we only need to invoke an off-the-shelf optimizer on the small set of prompts participating in that operation.} This sidesteps the difficulties of prior approaches simultaneously, since we never have to fit the full trajectory into the optimizer's context, propagate textual signals along a long causal chain, or replay the entire memory pipeline.

% \begin{figure}[t!]
% \centering
% \begin{subfigure}[t]{0.55\linewidth}
%   \centering
%   \includegraphics[width=\linewidth]{figures/optim_overview.pdf}
%   \caption{Pipeline}
%   \label{fig:optim_overview}
% \end{subfigure}\hfill
% \begin{subfigure}[t]{0.40\linewidth}
%   \centering
%   \includegraphics[width=\linewidth]{figures/optim_performance.pdf}
%   \caption{Performance}
%   \label{fig:optim_performance}
% \end{subfigure}
% \caption{{\bf Automatic optimization of Mem0.} 
% (a) Overview of the optimization pipeline. (b) Performance comparison of Mem0 before and after optimization, showing a 7.62\% improvement after three rounds.}
% \label{fig:mem0_optimization}
% \end{figure}

We evaluate this closed-loop pipeline on Mem0 and LoCoMo using an LLM-as-a-judge score. To isolate the effect of \memtrace, we introduce a no-attribution baseline. As shown in Figure~\ref{fig:optim_performance}, \memtrace-guided optimization improves performance from 59.08\% to 66.70\% ($+7.62$ points), whereas the no-attribution baseline decreases it to 44.73\%. 
% Without localization, the optimizer cannot determine how upstream extraction and memory-update prompts contribute to the final error, resulting in highly noisy updates. 
% \textbf{These results demonstrate the importance of localized credit assignment, even though \memtrace itself is imperfect (72.5\% OIA).} 
\textbf{These results demonstrate the importance of localized credit assignment: by identifying the operations responsible for failures, \memtrace provides meaningful insights that enable the optimizer to make effective updates, even though \memtrace itself is imperfect.}
Further experimental details and cost analysis are provided in Appendix~\ref{app:opt-details}.

% To evaluate this closed-loop optimization pipeline, we implement it on Mem0 as the target memory system and LoCoMo as the benchmark, using an LLM-as-a-judge score as the evaluation metric. We randomly sample three users from LoCoMo as the training split and reserve the remaining seven for testing. As shown in Figure~\ref{fig:optim_performance}, three rounds of optimization improve Mem0's performance by $7.62\%$ on the held-out test split. \textbf{Notably, this improvement is achieved despite the fact that \memtrace is not perfectly accurate (72.5\% operation identification accuracy), suggesting that even imperfect graph-based credit assignment can provide sufficiently useful optimization signals for practical prompt tuning.} Detailed experimental settings and the optimization cost breakdown are reported in Appendix~\ref{app:opt-details}.

%% file: section/7.related_work.tex
\section{Related Work}
\label{app:related}

Recent LLM memory systems aim to support long-horizon interactions by dynamically extracting, updating, forgetting, and maintaining memories across sessions \citep{Packer2023MemGPTTL, DBLP:conf/aaai/ZhongGGYW24, Xu2025AMEMAM, Chhikara2025Mem0BP, Cao2025RememberMR, Wang2025MIRIXMM, Fang2025LightMemLA, Liu2026SimpleMemEL, Hu2026EverMemOSAS, Yang2026PlugMemAT}. These systems introduce complex execution pipelines, making failures difficult to localize and attribute. Existing work on diagnosing LLM agents mainly focuses on identifying faulty steps in short reasoning traces within a single task instance, using sampling-based signals, process-level supervision, or LLM-based inspection of intermediate trajectories \citep{DBLP:conf/emnlp/XiongSZWWWLPL24, DBLP:conf/iclr/LightmanKBEBLLS24, DBLP:conf/acl/Wang0L025, Zhang2025AgenTracerWI, Ge2025WhoII, Baker2025MonitoringRM, DBLP:conf/icml/ZhangY0LHZL0W0W25, DBLP:journals/nature/YuksekgonulBBLLHGZ25, lee2026meta, li2026codetracer, Wang2026FromFL}. 
In contrast, failures in stateful agents with long-term memory may originate from earlier sessions and must be distinguished from substantial irrelevant interaction history. For more related work, see Appendix~\ref{app:more_related_work}. 

%% file: section/8.conclusion.tex
\section{Conclusion}
% We study the problem of tracing and attributing errors in LLM memory systems.
% We build \bench from public datasets and open-source memory systems, and propose \memtrace to attribute failures to concrete pipeline operations.

In this work, we study a new research question: how to automatically diagnose failures in non-parametric memory systems. 
To this end, we construct \bench, a benchmark built from publicly available datasets and open-source memory systems. Based on this benchmark, we propose \memtrace that attributes memory-system failures to concrete operations in the execution pipeline. Our findings show that precisely localizing failures remains challenging. We hope our work will encourage further research on automatic failure diagnosis for memory-augmented AI systems.

%% file: section/appendix.tex
\section{More Related Work}
\label{app:more_related_work}

\paragraph{Non-Parametric Memory Systems for LLMs.}
LLMs natively rely on the context window as a transient memory buffer, while KV caching \citep{DBLP:conf/mlsys/PopeDCDBHXAD23} avoids recomputing repeated prefixes. However, long-context prompting is fundamentally bounded by context length, computational cost, and well-known degradation phenomena such as lost-in-the-middle \citep{DBLP:journals/tacl/LiuLHPBPL24} and failures in long multi-turn conversation \citep{Laban2025LLMsGL}. Retrieval-Augmented Generation (RAG) externalizes memory into non-parametric stores, typically by chunking raw content and retrieving relevant units at inference time \citep{DBLP:conf/nips/LewisPPPKGKLYR020, DBLP:conf/acl/TrivediBKS23, DBLP:conf/iclr/AsaiWWSH24, DBLP:conf/emnlp/Li00MB24}. More recent RAG methods further improve retrieval by refining or summarizing raw content with LLMs before indexing \citep{DBLP:conf/iclr/SarthiATKGM24, DBLP:conf/nips/GutierrezS0Y024, Edge2024FromLT}. Beyond long-context prompting and RAG, recent memory systems focus on dynamically managing memories during open-ended interactions, including memory extraction \citep{Li2025MemOSAO, Fang2025LightMemLA, Hu2026EverMemOSAS, Liu2026SimpleMemEL}, updating \citep{Xu2025AMEMAM, Chhikara2025Mem0BP, Hu2026BeyondRF}, forgetting \citep{DBLP:conf/aaai/ZhongGGYW24, Cao2025RememberMR}, and multi-type memory maintenance \citep{Packer2023MemGPTTL, Wang2025MIRIXMM, Yang2026PlugMemAT}. Compared with the previous two paradigms, these systems involve substantially more complex execution pipelines, making their failures harder to localize and attribute.

\paragraph{Automatic Failure Attribution.}
Automatic failure attribution is studied in many domains, including software debugging \citep{DBLP:conf/esec/Zeller99, DBLP:conf/sigsoft/Zeller02}, cloud-service diagnosis \citep{Yu2021MicroRankEL, Zhang2025DynaCausalDC}, and deep learning analysis \citep{DBLP:conf/icml/SundararajanTY17, DBLP:conf/nips/MengBAB22}. The most related line of work focuses on localizing failures in LLM agent systems. Some methods identify faulty steps through sampling-based or process-level signals \citep{DBLP:conf/acl/Wang0L025, DBLP:conf/emnlp/XiongSZWWWLPL24, DBLP:conf/iclr/LightmanKBEBLLS24, Zhang2025AgenTracerWI, Ge2025WhoII}. 
Others use another LLM or agent to inspect intermediate traces and diagnose error locations \citep{Baker2025MonitoringRM, DBLP:conf/icml/ZhangY0LHZL0W0W25}. 
Some approaches further exploit structured trajectories such as trees \citep{lee2026meta, li2026codetracer} or graphs \citep{DBLP:journals/nature/YuksekgonulBBLLHGZ25, Wang2026FromFL} for failure localization. \citet{Qian2026TheWB} study failure attribution in memory-driven interaction scenarios, analyzing how retrieved memories influence an agent's downstream actions. This corresponds most closely to response-stage attribution in our work. Overall, prior work mainly targets short reasoning traces within a single task instance. By contrast, in stateful agents with non-parametric memory, failures may be introduced long before the final answer is produced, potentially in earlier sessions, and must be distinguished from substantial irrelevant interaction history.

\paragraph{Memory Benchmarks.} As memory systems develop rapidly, automatic evaluation for memory systems becomes increasingly important. LoCoMo \citep{DBLP:conf/acl/MaharanaLTBBF24} is one of the earliest and most widely used benchmarks for evaluating long-term memory in LLMs. Later work improves the difficulty of user trajectories \citep{DBLP:conf/iclr/WuWYZCY25, Bian2026RealMemBL}, increases their diversity \citep{Hu2025EvaluatingMI, DBLP:conf/acl/Tan000DD25}, or enriches them with multimodal information \citep{Wang2025MIRIXMM, Jiang2025PersonaMemv2TP, Bei2026MemGalleryBM}. Other benchmarks shift the evaluation focus to different scenarios \citep{DBLP:conf/iclr/Zhao00HL25, Du2025MemGuideIM, Hu2026CloneMemBL, He2026MemoryArenaBA}. These benchmarks usually construct question-answer pairs to measure the final performance of memory systems. However, this evaluation paradigm provides limited fine-grained diagnostic information. HaluMem \citep{Chen2025HaluMemEH} offers a more fine-grained automatic evaluation by assessing the accuracy of memory extraction and memory updating. However, it mainly checks whether target memories can be found in the current memory store through retrieval, which may not always reflect the true system behavior. It also cannot reveal when an error is introduced or which operation causes it. In contrast, our work only requires the golden answer, rather than manually provided source evidence or golden memories, to automatically perform fine-grained diagnosis of memory-system failures. Therefore, it can serve as diagnostic infrastructure for different memory benchmarks.

\section{Detailed Discussion of Prior Failure Attribution Work}
\label{app:prior_failure_attribution_discussion}

\memtrace is not the first approach to use execution traces for failure attribution. Its contribution is to study this paradigm in stateful agents with non-parametric memory, where persistent states evolve across sessions and a failure may originate from a much earlier operation relevant to memory construction. Below, we clarify how this setting relates to three groups of closely related work and why they do not provide directly applicable baselines.

\paragraph{Agent Failure Attribution.}
Prior works on agent failure attribution, such as AgenTracer \citep{Zhang2025AgenTracerWI}, CausalFlow \citep{Bonagiri2026CausalFlowCA}, and GraphTracer \citep{Zhang2025GraphTracerGF}, primarily study task-scoped agent trajectories. These methods use counterfactual intervention or replay to identify failure-inducing steps. \textbf{Directly applying this strategy to long-term memory systems is computationally difficult}: repairing an early memory operation requires reconstructing the memory state and replaying potentially hundreds of subsequent interactions.  Repeating this intervention for every candidate operation would be prohibitively expensive. Moreover, \textbf{replay does not perfectly instantiate the idealized intervention in our problem definition}, because errors in later operations may still prevent the final answer from being corrected after an early fault is repaired.

GraphTracer is particularly close to our work because it also represents execution traces as dependency graphs. However, it constructs these graphs by incrementally converting unstructured agent trajectories into structured dependencies using LLMs and pattern matching, followed by consistency checking. \textbf{This construction becomes expensive for long trajectories containing many upstream operations or agents, and dependencies inferred from textual traces are not guaranteed to reproduce the program's actual runtime information flow.} In contrast, \trace uses lightweight source-code instrumentation to record operation-variable dependencies during execution. It directly captures the lifecycle of persistent memory variables, including creation, in-place update, deletion, retrieval, and downstream use, without reconstructing dependencies from unstructured logs. 
Adapting GraphTracer to our setting would require both representing versioned memory states and scaling graph construction to multi-session traces containing millions of tokens. A promising future direction is to use LLMs to insert lightweight instrumentation automatically and then obtain runtime-faithful execution graphs by running the instrumented programs.

\paragraph{RAG Diagnostics.}
Prior works on RAG diagnostics, such as RAGChecker \citep{DBLP:conf/nips/RuQHZSCJWSLZWJ024} and RAGAs \citep{DBLP:conf/eacl/ESJAS24}, evaluate retrieval and generation quality through component-level metrics computed from the final retrieved context and generated response. \memtrace differs from these methods in its prediction target: rather than reporting component-level quality, it identifies a concrete faulty operation and explains why that operation causes the observed failure. For example, within EverMemOS, an incorrect sufficiency judgment, a failure in a subsequent retrieval step, and a reranking failure may all manifest as low context recall, but \memtrace distinguishes these underlying causes and points to the specific operation that requires correction. 
Moreover, unlike RAG-specific diagnostic methods, \memtrace is not restricted to RAG pipelines and can be applied to different memory-system architectures, including failures introduced during memory construction.

\paragraph{Operation-Level Memory Evaluation.}
Prior works such as HaluMem \citep{Chen2025HaluMemEH} evaluate extraction, update, and deletion operations against annotated golden memory points. This proactive evaluation is effective when intermediate supervision is available. \memtrace addresses a complementary setting: it starts from an observed end-task failure and traces backward through the actual execution graph using the question and reference answer, without requiring a golden memory state for every session or validating every memory operation in advance. It can also attribute failures in complex retrieval or response pipelines beyond predefined memory-maintenance operations.

\section{Non-Parametric Memory Systems}
\label{app:memory_system_definition}

Consider a historical observation trajectory of length $n$, denoted as $\tau = \{m_i\}_{i=1}^{n}$, and an associated question-answer pair $(q, a)$. Here, $m_i$ denotes the $i$-th message in the trajectory, $q$ is a question whose answer requires certain critical information from the historical observations, and $a$ is the corresponding golden answer. We assume that the timestamp of the question satisfies $t_q > t_{m_n}$, meaning that the question is asked after all messages in the trajectory have
been observed. Let $\mathcal{M}$ denote a non-parametric memory system, $\mathcal{U}_{\mathcal{M}}$ denote the memory update operation, and $\mathcal{R}_{\mathcal{M}}$ denote the memory read operation. During memory construction, the memory system incrementally updates its memory state in a message-by-message manner: 
\begin{equation}
\mathcal{S}_{j} = \mathcal{U}_{\mathcal{M}}(\mathcal{S}_{j-1}, m_j),
\quad 1 \leq j \leq n .
\end{equation}

Given the question $q$ at time $t_q$, the system reads relevant context $\mathcal{C}$ from the latest memory state $\mathcal{S}_n$ and generates an answer $\hat{a}$ based on the question-answering model $\mathcal{Q}$:
\begin{equation}
\begin{aligned}
\mathcal{C} &= \mathcal{R}_{\mathcal{M}}(\mathcal{S}_n, q), \\
\hat{a} &= \mathcal{Q}(q, \mathcal{C}).
\end{aligned}
\end{equation}

Under this formulation, both long-context models and RAG systems can be viewed as instances of non-parametric memory systems. For long-context models, the update operation simply appends each new message to the context, and the read operation returns the entire memory state as the input context. For RAG systems, the update operation typically stores raw
messages in a vector database, while the read operation performs top-$K$ semantic retrieval. More advanced memory systems further introduce large language models into either the memory update operation, the memory read operation, or both. This increasing complexity introduces many sub-operations into memory updates and reads, making it difficult to find the source of errors.

% Main overview figure with clickable thumbnails
\begin{figure*}[t!]
\centering
\hypertarget{fig:annotation-interface-overview-target}{}

\begin{minipage}[t]{0.3\linewidth}
\centering
\hyperref[fig:annotation-interface-left]{%
  \includegraphics[width=\linewidth]{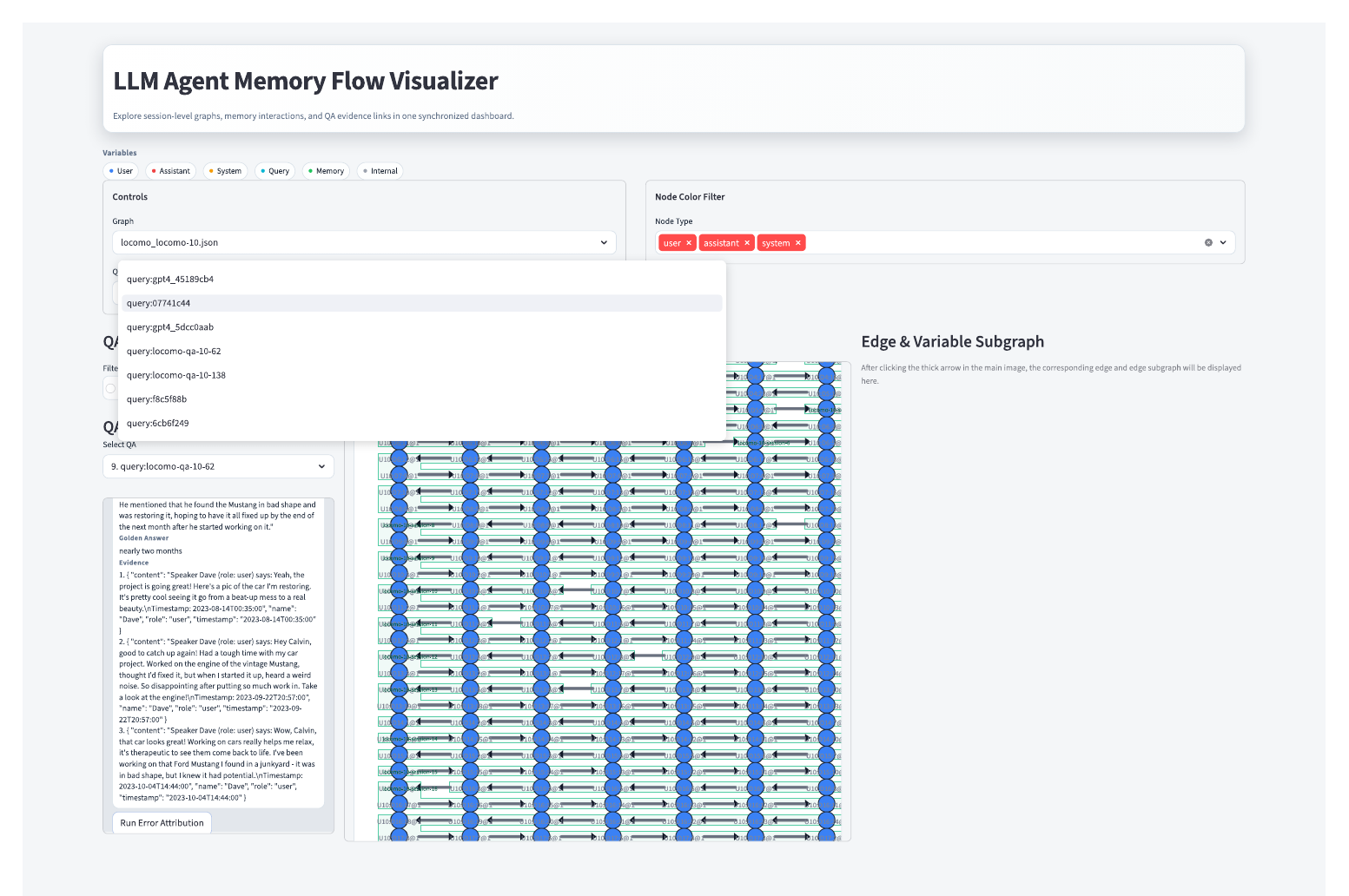}%
}
\end{minipage}
\hspace{0.01\linewidth}
\begin{minipage}[t]{0.2725\linewidth}
\centering
\hyperref[fig:annotation-interface-middle]{%
  \includegraphics[width=\linewidth]{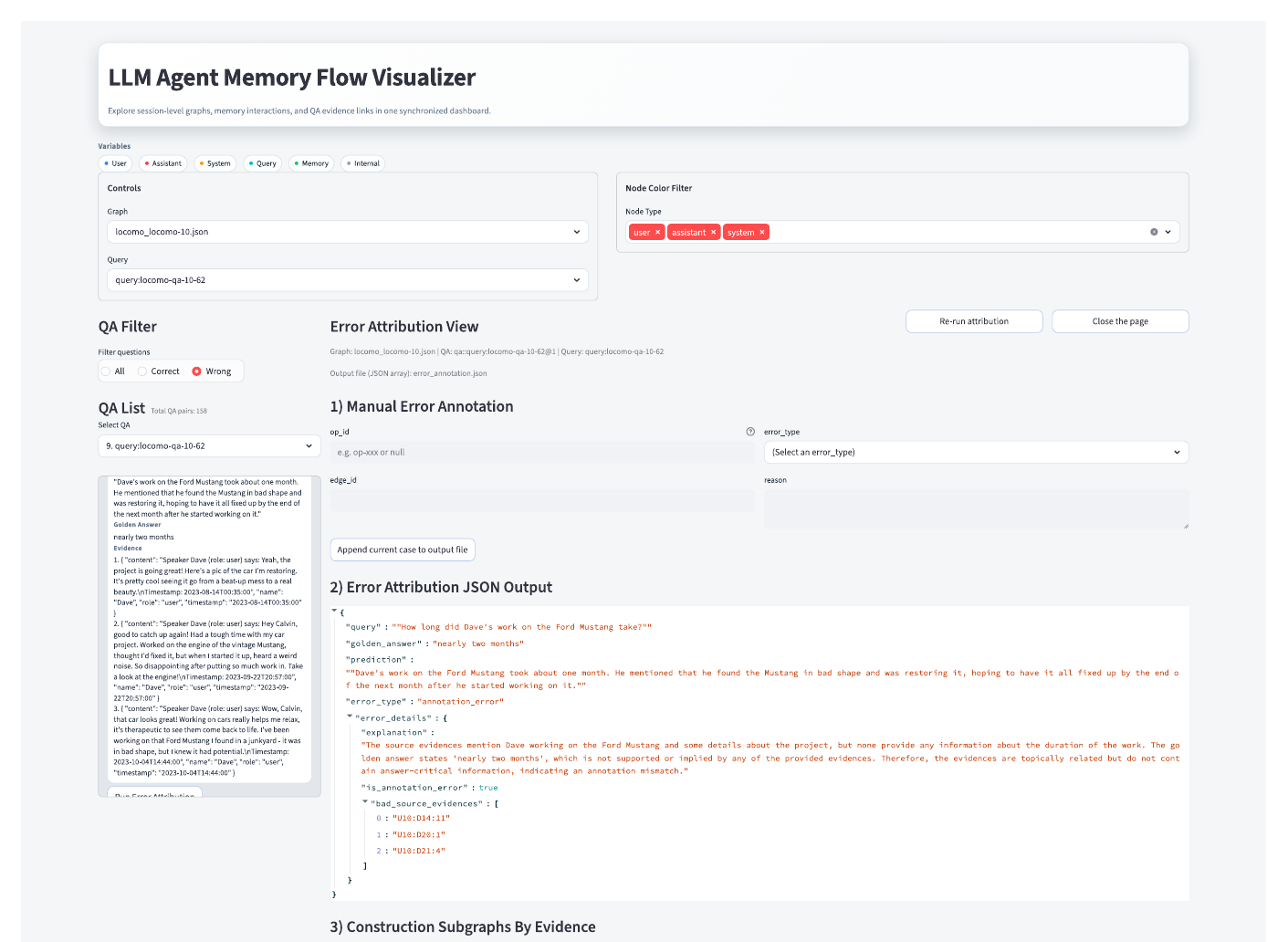}%
}
\end{minipage}
\hspace{0.01\linewidth}
\begin{minipage}[t]{0.374\linewidth}
\centering
\hyperref[fig:annotation-interface-right]{%
  \includegraphics[width=\linewidth]{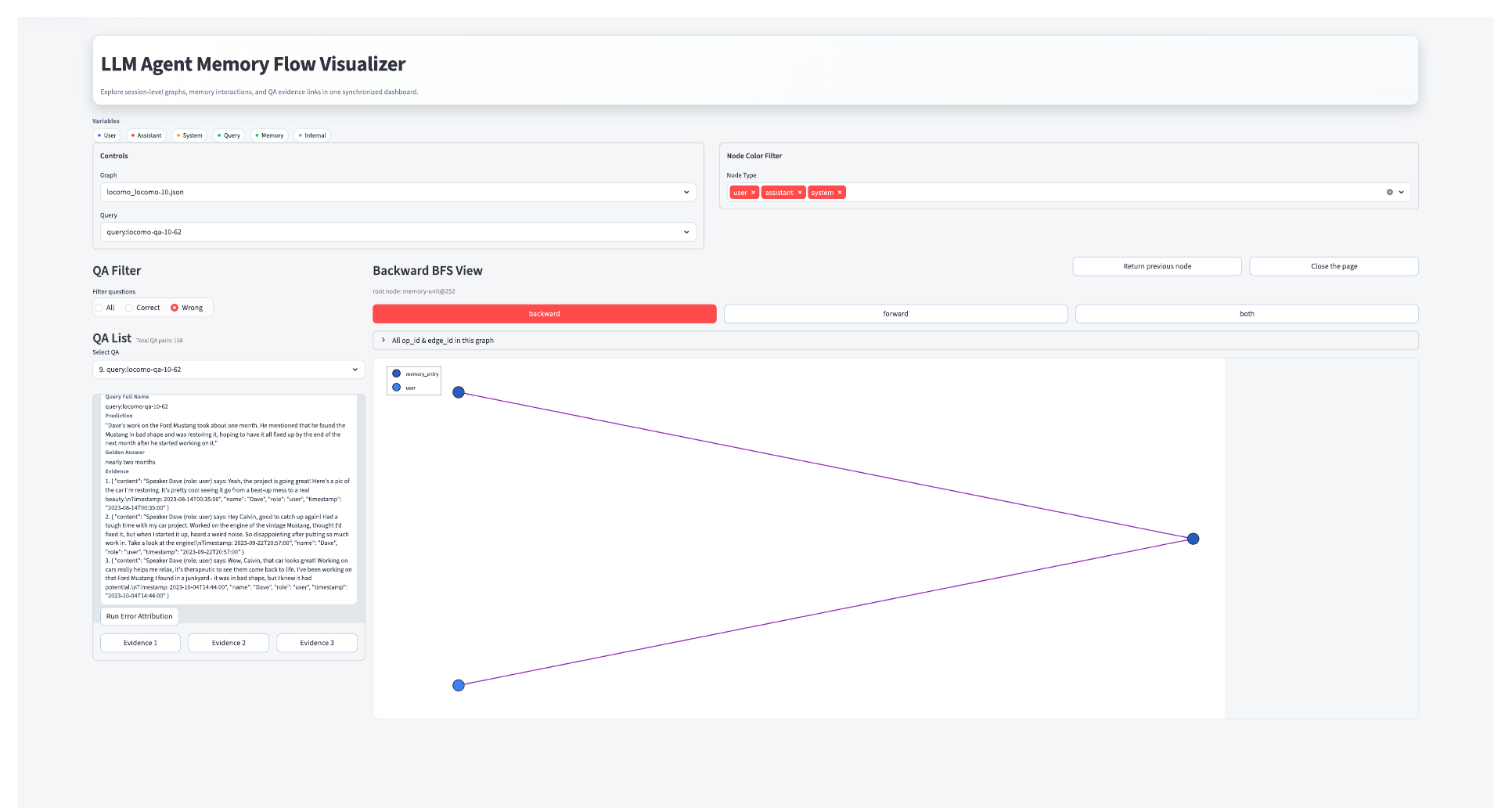}%
}
\end{minipage}

\caption{\textbf{The annotation interface with each thumbnail clickable and linking to its corresponding full-size visualization.} The left panel shows the entry point to the annotation interface, with the full-size visualization shown in Figure~\ref{fig:annotation-interface-left}. The middle panel presents the annotation submission view, with the full-size visualization shown in Figure~\ref{fig:annotation-interface-middle}. The right panel provides an interface for exploring the execution graph, with the full-size visualization shown in Figure~\ref{fig:annotation-interface-right}.}
\label{fig:annotation-interface-overview}
\end{figure*}

\section{Dataset Construction}
\label{app:dataset_construction}

\subsection{Data Sources and Memory Systems}
\label{app:data_sources} 

To obtain diverse interaction settings, we collect question-answer pairs from LoCoMo \citep{DBLP:conf/acl/MaharanaLTBBF24}, LongMemEval \citep{DBLP:conf/iclr/WuWYZCY25}, and RealMem \citep{Bian2026RealMemBL}. These datasets differ substantially in their trajectory structure and evaluation protocols. In LoCoMo and LongMemEval, questions are posed after the interaction trajectory has been completed. LoCoMo provides multiple questions for each trajectory, whereas LongMemEval provides a single question per trajectory. By contrast, RealMem interleaves questions within the interaction, where each task query corresponds to a user message and the gold answer is the subsequent assistant response. 

Accordingly, we use different evaluation protocols for these datasets when collecting execution graphs. For LoCoMo and LongMemEval, we adopt an offline protocol, where the full interaction trajectory is first processed by the memory system and all questions are evaluated afterward. For RealMem, we follow an online protocol. When a user message corresponds to a task query, the system answers the question using only the memory state available at that point, before the message is incorporated into memory.

We select four representative memory systems, namely long-context memory, RAG, Mem0, and EverMemOS. These systems cover a broad range of memory construction and retrieval mechanisms. The long-context baseline maintains the entire interaction history as a single memory unit and updates it with a rule-based procedure. RAG constructs an external memory store without any information loss. Mem0 supports not only memory addition but also memory update and deletion operations. EverMemOS uses a more complex retrieval pipeline involving LLM-based sufficiency judgment and query refinement. 

Together, the heterogeneous source datasets and diverse memory systems provide a strong basis for collecting execution graphs that vary in both structural properties and failure modes.

\subsection{Execution Graph Collection Details}
\label{app:collection_details}

To collect each memory system's execution graphs, we instrument each memory system's source code with \trace tracing statements. During instrumentation, we carefully choose the granularity of operations to make failure attribution meaningful and reduce annotation ambiguity. In some cases, we merge multiple low-level operations into a single traced operation when their failures are difficult to separate reliably. For example, in EverMemOS, retrieval involves an initial search, an LLM-based sufficiency judgment, optional query refinement, a second search, and result aggregation. If the system fails because key memory units are missing while the LLM judges the retrieved evidence as sufficient, it is ambiguous whether the error should be attributed to the initial retrieval or to the sufficiency judgment. Merging these tightly coupled steps into one traced operation avoids such borderline cases.

After instrumentation, we run all four memory systems on a sampled set of trajectories, including 4 from LoCoMo, 200 from LongMemEval, and 3 from RealMem. Each memory system processes each interaction trajectory message by message and produces one execution graph per trajectory. The detailed system configurations are provided in Appendix~\ref{app:memory_system_setup}. In total, we collect 1,514 distinct errors across all systems after filtering out unanswerable questions\footnote{We currently only consider questions with available source evidence. Unanswerable questions, by definition, do not have such evidence.}.  Among them, RAG, long-context, Mem0, and EverMemOS account for 467, 279, 466, and 302 errors, respectively. The corresponding numbers of execution graphs are 138, 132, 87, and 80.

\subsection{Experimental Setup for Memory Systems}
\label{app:memory_system_setup}

We standardize the backbone models and evaluation settings across all memory systems to ensure fair comparison, particularly for subsequent failure mode analysis. Specifically, both the memory construction model and the downstream response generation model are set to GPT-4.1 mini \citep{GPT41}.  For dense retrieval, we use Qwen3-Embedding-4B \citep{Zhang2025Qwen3EA} as the embedding model.  For EverMemOS, the reranker is Qwen3-Reranker-4B \citep{Zhang2025Qwen3EA}.  Both models are deployed using vLLM \citep{DBLP:conf/sosp/KwonLZ0ZY0ZS23}.  The number of retrieved memory units is fixed to 10 for all systems.

For evaluation, we adopt an LLM-as-a-Judge protocol \citep{Gu2024ASO} across all datasets, with dataset-specific prompts.  For LoCoMo, the evaluation prompt is adapted from the Mem0 paper. 
For LongMemEval, we follow the official implementation. For RealMem, we adapt the official prompt by converting the original scoring scheme from a 0-3 scale to a binary format. The judge model is Claude Opus 4.5 \citep{CLAUDE45}, which helps mitigate evaluation bias due to architectural differences from GPT-based models \citep{DBLP:conf/nips/ZhengC00WZL0LXZ23, DBLP:conf/acl/LiuML24}.

For the long-context baseline, the context window is set to 32k tokens. For the RAG baseline, we adopt an online chunking strategy to construct memory units. Concretely, we maintain a message buffer, and when adding a new message would cause the total token count to exceed 1200, 
the existing buffer (if non-empty) is flushed before inserting the new message. The flushed messages are concatenated to form a memory unit, which is then added in the memory store. For Mem0, we use its open-source implementation without graph-based extensions, and retain the default prompts and configurations. For EverMemOS, we adapt the official evaluation pipeline. 
For boundary detection, the token limit and message limit are set to 8192 and 50, respectively. 
During hybrid retrieval, both sparse and dense retrievers return top-50 memory units. After the initial hybrid retrieval and reranking, the top 10 memory units are used for sufficiency checking. Multiple refined queries may be generated for the second-stage retrieval. For reciprocal rank fusion (RRF) in EverMemOS, we set the fusion constant to 60.

\subsection{Annotation Process}
\label{app:annotation_process}

We recruit five annotators from the author team to label failure information. All annotators have substantial experience with LLM agents and are familiar with the memory construction and retrieval mechanisms of the four target memory systems. Before the annotation process, we define an error taxonomy (Appendix~\ref{app:error_taxonomy}) and provide detailed annotation guidelines spanning more than 10 pages\footnote{Annotators are allowed to use their knowledge of each memory system to simplify annotation. For example, RAG does not introduce memory-construction errors in our setup, and long-context memory does not introduce retrieval errors because all retained context is provided directly to the response generator.}. Given that each execution graph may contain thousands of nodes, direct inspection is challenging. To facilitate efficient and reliable annotation, we develop an interactive annotation interface (see Appendix~\ref{app:annotation_interface}).

The annotation process consists of three stages. In the first stage, for each memory system, we randomly shuffle the collected erroneous cases and present them to annotators in a consistent order. Annotators then review the cases sequentially and stop once 40 system-related errors have been identified\footnote{Errors arising from annotation mistakes or LLM-as-a-judge issues are excluded from this quota, as they are not attributable to the system itself. 
Nevertheless, such errors are still recorded without an upper bound.}. 
Each memory system is assigned to three annotators, who work independently without communication during this stage. 
For each erroneous case, annotators are required to specify the error type, identify the earliest faulty operation, and provide a natural-language explanation of the failure. 
After the first stage, a small fraction of cases may still have fewer than three annotations. 
In the second stage, we identify these cases and notify each annotator of the remaining cases they need to label, ensuring that every annotated case receives labels from three annotators. 
In the third stage, for cases whose final label cannot be determined by majority voting, the annotators discuss the case collectively.
During this discussion, we require annotators to revisit the execution graph and failure explanation until a consensus label is reached. Ultimately, this process yields 160 system-related failure cases, along with 67 annotation errors and 3 LLM-as-a-Judge errors.

\subsection{Error Taxonomy}
\label{app:error_taxonomy}

According to the lifecycle of information flow in stateful agents with non-parametric long-term memory, we define seven error types as follows.

\paragraph{Annotation Error.} The source-evidence set associated with a QA pair is insufficient to support the reference answer, or the reference answer itself is incorrect with respect to the evidence.

\paragraph{LLM-as-a-Judge Error.} The system's answer is in fact acceptable, but the automatic judge incorrectly marks it as wrong.

\paragraph{Extraction Error.} The critical information is never captured into any memory unit during memory construction, and thus never enters the memory store.

\paragraph{Update Error.} A memory unit initially contains the critical information, but a subsequent update operation modifies the unit in a way that removes or degrades that information.

\paragraph{Deletion Error.} A memory unit containing the critical information exists but is explicitly removed (e.g., due to memory management), resulting in irreversible information loss.

\paragraph{Retrieval Error.} At retrieval time, the memory store does contain the required information, but the retrieval pipeline fails to include it in the final retrieved context.

\paragraph{Response Error.} The retrieved context contains all necessary evidence, yet the final LLM still produces an incorrect answer.

\subsection{Annotation Interface Design}
\label{app:annotation_interface}

Figure~\ref{fig:annotation-interface-overview} illustrates our annotation interface. Due to the large scale of execution graphs (averaging 2,262.65 nodes and 3,613.70 edges), it is impractical to display the entire graph at once. Therefore, the interface is designed to present only partial views tailored to the annotation workflow.

Since memory construction operations occur between pairs of messages, the entry panel focuses on explicit interactions between the user and the AI assistant. The lower-left section displays the current question-answer pair, including the question, golden answer, model response, and associated source evidence. Several control buttons are provided. One triggers an auxiliary error attribution algorithm, while others highlight the positions of corresponding source evidence within the message stream. The control panel in the upper-left corner shows the graph file associated with the current question-answer pair, along with a shuffled list of erroneous cases. Annotators are required to follow this predefined order during annotation. 

Upon clicking the ``Run Error Attribution'' button, the system executes a preliminary automatic error attribution algorithm and transitions to the middle panel of Figure~\ref{fig:annotation-interface-overview}, which serves as the annotation submission interface. This panel presents the predicted error types and the locations of suspected faulty operations. These results are intended as guidance rather than definitive labels. In this interface, annotators can inspect multiple subgraphs to facilitate error localization. Specifically, they can examine (i) memory construction subgraphs induced by each source evidence\footnote{If a source evidence leads to the creation of a new memory unit, subsequent updates or deletions of that unit caused by other source evidence are not shown in the induced subgraph. Such downstream changes can be inspected in the execution graph exploration interface.}, (ii) retrieval subgraphs, and (iii) response generation subgraphs. If further inspection is needed, annotators can click on any variable within these subgraphs to navigate to the execution graph exploration interface (the right panel in Figure~\ref{fig:annotation-interface-overview}). This interface supports both backward tracing (identifying the variables that produce the current variable) and forward tracing (tracking its downstream effects), enabling flexible exploration of the execution graph.

Overall, our annotation interface significantly improves annotation efficiency by structuring complex execution graphs into manageable, task-oriented views.

% \subsection{Singleton Decisive Error Sets in Sequential Memory Systems}
% \label{app:singleton_error_set}

% The formal definition in Section~\ref{sec:problem_formulation} allows the decisive error set \(\mathcal{O}^*\) to contain multiple faulty operations, which is necessary for systems with concurrent or asynchronous execution. In our benchmark, however, all four target memory systems execute operations strictly sequentially. This also holds for RAG  in our implementation, where the chunker operates online and maintains a single message buffer. Since the execution is strictly sequential, the execution induces a total order over operations, implying that there exists a unique earliest faulty operation whose correction suffices to successfully rescue the failed trajectory. Therefore, \(\mathcal{O}^*\) degenerates to a singleton consisting of this operation.

% Based on this property, annotators only need to identify one operation identifier for each labeled example, together with the error type and a natural-language explanation. 

\section{Tracing Toolkit}
\label{app:Tracing Toolkit}
\trace is a lightweight tracing package for recording developer-specified operations and the variables flowing through them. 
It is designed to collect execution graphs from existing Python systems without requiring developers to rewrite the original implementation around a new runtime abstraction. 

\subsection{Comparison with Prior Tracing Frameworks}
\label{app:tracing_framework_comparison}

A wide range of open-source frameworks are developed for collecting execution traces. Broadly speaking, these frameworks follow two design philosophies. The first philosophy is \emph{instrumentation-based tracing}, adopted by systems such as MLflow \citep{Zaharia_Accelerating_the_Machine_2018}, VizTracer \citep{Gao_VizTracer_-_A}, PySnooper \citep{rachum2019pysnooper}, and Langfuse \citep{Langfuse2026}. These frameworks ask developers to insert decorators, context managers, logging calls, or other hook-like statements at selected locations in the source code, in order to capture execution events such as function inputs, outputs, and auxiliary metadata. A major advantage of this design is that it is largely non-intrusive to the underlying application logic. Developers usually do not need to rewrite the data schema or reorganize the program around a new runtime abstraction. However, such frameworks are primarily event-centric. They can record spans, function calls, and metadata, but they generally do not make dependencies among intermediate variables first-class objects. As a result, they are less suitable for questions that require tracing how a particular memory unit is produced, how it is later modified, and which earlier variables causally contributed to a downstream failure.

The second philosophy is \emph{abstraction-native tracing}, where computation is expressed through framework-specific data containers or operators that can be automatically intercepted and organized into graphs. Representative examples include TensorFlow \citep{DBLP:conf/osdi/AbadiBCCDDDGIIK16}, PyTorch \citep{DBLP:conf/nips/PaszkeGMLBCKLGA19}, TextGrad \citep{DBLP:journals/nature/YuksekgonulBBLLHGZ25}, Trace \citep{DBLP:conf/nips/ChengNS24}, and DSPy \citep{khattab2023dspy}. These frameworks can often capture execution traces automatically and, in some cases, construct computation or dataflow graphs that encode dependencies among intermediate variables and operations. Their strength, however, comes from a strong assumption that the program must be written in terms of the framework's own abstractions. This is effective when the target domain already admits a stable computational substrate, such as tensors, modules, or other framework-defined objects. 
It is much less suitable for memory systems, whose schemas are heterogeneous, whose operations are highly flexible, and whose key entities (e.g., memory units, retrieval results, summaries, updates, and prompts) do not naturally fit into a single predefined abstraction.

These limitations motivate us to develop \trace. Conceptually, \trace combines the flexibility of instrumentation-based tracing with the provenance benefits of graph-based tracing. It uses explicit instrumentation, but instead of recording only events or call trees, it allows developers to trace arbitrary Python variables through user-defined serializable representations and to explicitly record dependencies among variables and operations. Both variables and operations can be annotated with comments and semantic metadata, which makes the resulting execution graph easier to inspect and interpret. In addition, \trace supports in-place updates through versioned variable nodes, allowing us to recover the evolution trajectory of a memory unit over time rather than only its final state.

This design is particularly useful for stateful agents with non-parametric memory, where one often needs to inspect not only what operation occurs, but also how a memory artifact is created, updated, retrieved, and eventually used in producing an answer. More broadly, \trace is not specific to memory systems. It can also be applied to other programs with rich evolving state, such as dynamic task planning or business data workflows. We view it as a general-purpose toolkit for collecting execution graphs that can support future research on automatic failure attribution and program understanding.

At the same time, \trace inherits the main trade-off of explicit instrumentation. Because the graph is constructed through developer-authored tracing statements, its quality depends on instrumentation coverage and granularity. Unlike framework-native tracers, it does not automatically capture all computations by default, and developers must define suitable representations and identities for the variables they wish to track. In this sense, \trace prioritizes flexibility and semantic expressiveness over full automation. This design choice is motivated by a broader shift in modern AI-assisted development environments, 
where the cost of writing and modifying code is significantly reduced. As a result, the traditional 
trade-off between ease-of-use and flexibility is shifting, increasingly favoring expressivity and 
composability over rigid, abstraction-heavy designs.

\subsection{Design and Features of \trace}
\label{app:smartcomment_features}

\trace has the following key features.

\paragraph{Hierarchical Data Model.} \trace organizes traces using a hierarchical data model. At the top level, an execution graph stores the traced state of a program run. A graph contains sessions, which can be used to represent different stages of the system lifecycle, such as memory construction. Within each session, operations represent developer-specified computational steps, and variables represent the intermediate artifacts consumed or produced by these operations. Dependencies among variables are represented by edges associated with the corresponding operation. This design allows \trace to capture not only which operations are executed, but also how information flows across variables over time.

\paragraph{Explicit Instrumentation.} Execution graphs are built through lightweight instrumentation in \trace \footnote{In the engineering implementation, the internal graph stored by \trace is not strictly a bipartite graph in the sense of Section~\ref{sec:problem_formulation}. Variables are connected directly by edges, and each edge stores the identifier of the operation that induces the dependency.}. Developers only need to insert a small number of tracing statements at key locations to mark operation boundaries, inputs, outputs, and dependencies. This makes it possible to trace existing memory-system implementations without restructuring their code. To recognize the same variable as it is passed, copied, or updated across different parts of the program, \trace uses a global tracing context and user-definable identity functions. They help determine when two runtime objects should be treated as the same traced variable. This is especially important for memory systems, where the same memory unit may be passed through multiple components, updated in place, or re-created from serialized representations.

\paragraph{Rich Contextual Attributes.} \trace supports providing contextual information for sessions, operations, variables, and edges. Developers can assign category labels, natural-language comments, and custom metadata to each traced object. For example, for operations involving LLM calls, the metadata can record the model name, hyperparameters for text generation, or error messages returned by the API. These contextual attributes provide semantic context beyond raw execution events, making the resulting graphs easier for both humans and agents to inspect, interpret, and debug.

\paragraph{Persistence and Visualization.} \trace supports graph export and import, enabling execution graphs to be persisted across program runs and restored for later analysis. It also provides visualization utilities based on PyVis\footnote{https://github.com/westhealth/pyvis} and Graphviz\footnote{https://github.com/xflr6/graphviz}, which allow developers and annotators to inspect execution graphs interactively or as static diagrams. These capabilities are useful for validating instrumentation quality, and understanding complex system behavior.

\subsection{Instrumentation Example}
\label{app:smartcomment_example}

\begin{figure*}[t!]
\centering
\vfill 
\resizebox{1.0\textwidth}{!}{
\includegraphics{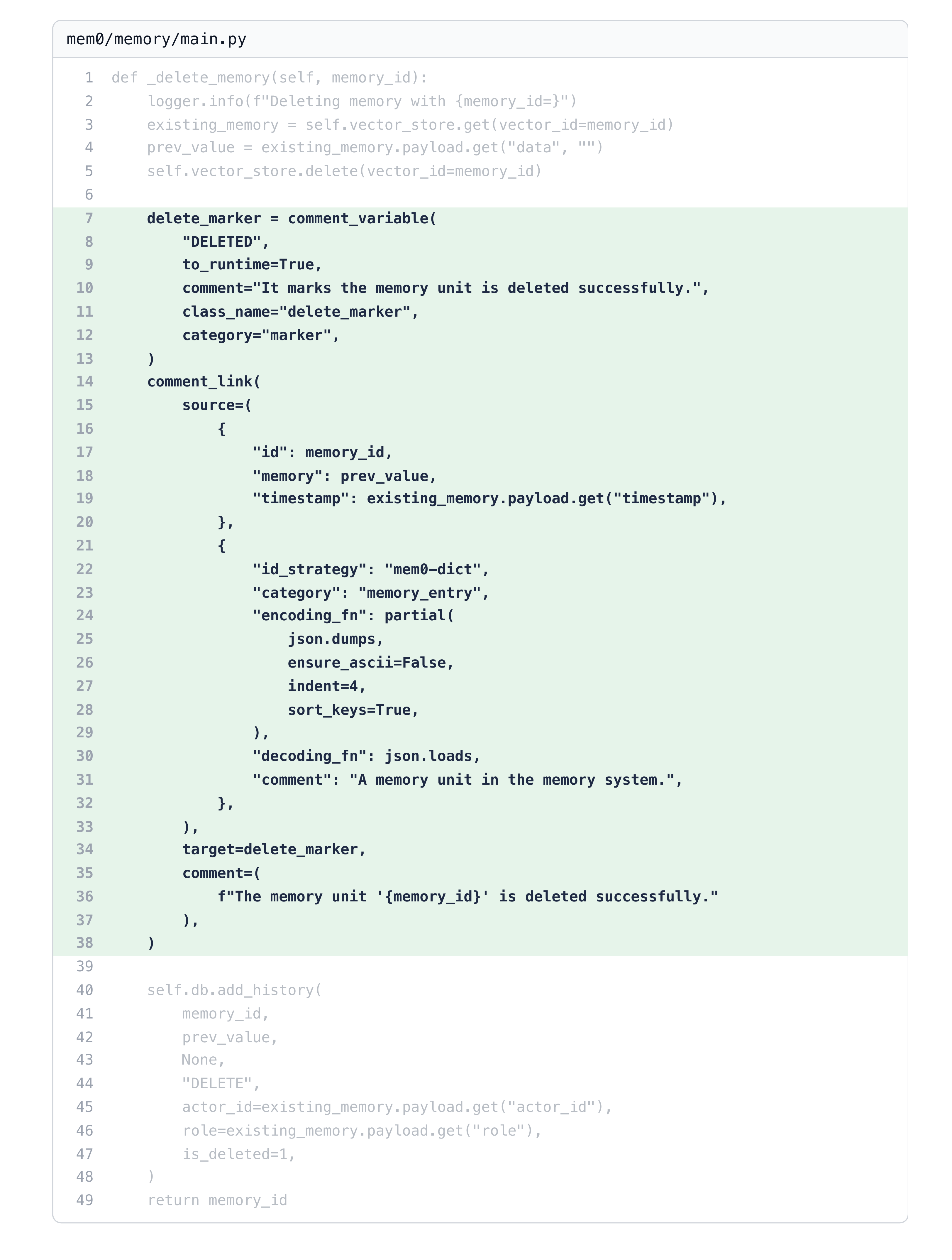}}
\caption{
\textbf{An instrumentation example.}  We insert two \trace statements, highlighted in green, into the method \texttt{\_delete\_memory} of the class \texttt{Memory} in the Mem0 source code to record the deletion of a memory unit. The method is extracted for presentation, with surrounding code omitted and indentation adjusted for readability.
}
\vfill 
\label{fig:instrumentation_example}
\end{figure*}

Figure~\ref{fig:instrumentation_example} shows an instrumentation example in Mem0\footnote{The full instrumentation details for all four memory systems are available in the released source code.}. We insert two \trace statements into the memory-deletion method of the class \texttt{Memory} to record the deletion of a memory unit. Since deletion removes the original memory unit from the memory store, we introduce a constant deletion marker and register it as a traced variable using \texttt{comment\_variable}. We then use \texttt{comment\_link} to connect the deleted memory unit to this marker, explicitly representing the deletion effect in the execution trace.

The \texttt{comment\_link} call is executed within the current operation context, so the resulting dependency edge is automatically associated with the corresponding operation identifier. In this example, the \texttt{source} argument of \texttt{comment\_link} is specified as a Python tuple. The first element is a snapshot of the deleted memory unit represented as a Python dictionary. The second element provides tracing configurations and contextual attributes of this memory unit. The identity strategy used here is registered in \trace as \texttt{mem0-dict}.

\section{Multi-Error Attribution}
\label{app:multi_error_attribution}

\paragraph{Background.}
The problem formulation in Section~\ref{sec:problem_formulation} focuses on identifying a single earliest decisive faulty operation. 
In more complex systems, however, a failure may be jointly caused by multiple independent errors, such that correcting any one of them alone is insufficient. For example, a multi-user memory system may maintain separate memory stores for different users while answering a question requires combining information across several users' histories. Independent errors in these stores can then jointly cause the final failure.

\paragraph{Decisive Error Set.}
The execution-graph formulation can represent these cases through a \textbf{Decisive Error Set}. 
Let \(O_c \subseteq \mathcal{O}\) be a candidate set of faulty operations. We say that \(O_c\) is valid if every operation in \(O_c\) is faulty, all operations in its strictly upstream ancestor set \(\mathrm{Anc}_{\mathcal{G}}(O_c)\) are functionally correct, and \(O_c\) is causally sufficient. For the last condition, we construct \(\mathcal{G}^{(O_c, *)}\) by replacing the faulty output variables of all operations in \(O_c\) with their correct counterparts and assuming ideal execution for all operations in \(\mathrm{Desc}_{\mathcal{G}}(O_c)\). The candidate set is causally sufficient if this intervention rescues the execution, i.e., \(Z(\mathcal{G}^{(O_c, *)})=0\). Let \(\mathcal{F}(\mathcal{G})\) contain all candidate sets satisfying these conditions. 
The decisive error set $\mathcal{O}^*$ is then defined by imposing a minimality constraint over this feasible space: removing any operation from $\mathcal{O}^*$ breaks causal sufficiency. This is expressed mathematically as:
\begin{equation}
\begin{aligned}
\mathcal{O}^* \in
\bigl\{
O_c \in \mathcal{F}(\mathcal{G})
\mid
&\nexists O' \in \mathcal{F}(\mathcal{G}) \\
&\text{s.t. } O' \subset O_c
\bigr\}.
\end{aligned}
\end{equation}
This formulation reduces multi-error attribution to identifying a minimal topological frontier of faulty operations that cause the system failure.

\paragraph{Extending \memtrace.}
Although current implementation of \memtrace targets singleton decisive error sets, the same framework can be extended to non-singleton cases. Once the agent identifies a decisive faulty operation, it can record it and prune operations strictly downstream of it, since these operations cannot belong to the same earliest decisive error set. The agent can then continue from the remaining to-explore list to search for additional independent faulty operations in the pruned execution graph. We leave this to future work.

\section{Additional Experimental Details for Failure Attribution}
\label{app:failure_attribution_details}

For both \memtracels and \memtrace, we set the working-context safety threshold to $T=272{,}000$ tokens and the maximum reasoning budget to 200 iterations. We use Qwen3-Embedding-4B \citep{Zhang2025Qwen3EA} as the embedding model used to initialize starting points. For \memtrace, the maximum to-explore list size is $N = 16$. For All-at-Once, we convert the execution graph into a weakly structured operation log. Because these logs are extremely long, we retain the latest 600,000 tokens and truncate earlier operations.

We also explore a long-context baseline inspired by MemAgent \citep{Yu2025MemAgentRL}. It divides the operation log into chunks and processes them sequentially. After each chunk, the agent updates its working memory. After processing the complete log, it predicts the earliest decisive faulty operation using the final working memory. However, this approach must scan the entire trajectory, including many failure-irrelevant operations. It is therefore prohibitively slow and performs poorly, so we exclude it from the comparison.

\section{Report Generation Details}
\label{app:report_generation}

We use GPT-5.4 to synthesize the error attribution results into a coherent error analysis report. The attribution results are obtained using different diagnostic settings for the two systems. For Mem0, we use the standard \memtrace setting. For EverMemOS, we leverage source evidence and prior knowledge. Specifically, we feed the model mini-batches of attribution outputs (with a batch size of four) and prompt it to iteratively update and refine the current report. When necessary, the model is also encouraged to identify finer-grained subtypes within each major error category. To improve the clarity and writing quality of the generated analysis, we include an exemplar error analysis section from the MMMU paper \citep{DBLP:conf/cvpr/YueNZ0LZSJRSWYY24} in the prompt as an in-context example.

\begin{wraptable}{r}{0.48\columnwidth}
\centering
\small
\begin{tabular}{lcc}
\toprule
Stage & Tokens & Time \\
\midrule
\memtrace & 493.21 & 1.33 \\
Feedback Generation & 17.19 & 0.23 \\
Feedback Aggregation & 16.25 & 1.23 \\
Prompt Update & 12.26 & 1.13 \\
\bottomrule
\end{tabular}
\caption{
Average cost of the closed-loop prompt optimization pipeline.
``Tokens'' denotes the average token cost, in thousands, including both
input and output tokens. ``Time'' denotes the average end-to-end runtime in minutes. For \memtrace and feedback generation, averages are computed per failed case. For feedback aggregation and prompt update, averages are computed per target prompt variable.
}
\label{tab:prompt_optimization_cost}
\end{wraptable}

\section{Prompt Optimization}

\subsection{Prior Prompt Optimization Methods} 
\label{app:prompt-opt}

Multi-session memory systems create very long chains between a prompt and the eventual failure it causes. For example, a fact extracted incorrectly in an early session may not lead to a visible mistake until hundreds of turns later. Before the error finally appears, the incorrect information may already have passed through many operations including memory update and memory retrieval. This property makes existing prompt optimization methods difficult to apply effectively.

\paragraph{Reflection-based optimizers.} Methods such as ACE \citep{Zhang2025AgenticCE}, and GEPA \citep{Agrawal2025GEPARP} feed the entire execution trajectory to an optimizer model. The optimizer is asked to reflect on the current performance and rewrite the prompts.  In multi-session memory systems, the resulting trajectory exceeds the optimizer's context window. Even when the trajectory fits, optimizer attention degrades on long inputs, so reasoning over the relevant operations remains unreliable \citep{DBLP:journals/tacl/LiuLHPBPL24, DBLP:conf/icml/ModarressiDDBR025}.

\paragraph{Candidate-and-replay search.} Methods such as OPRO \citep{DBLP:conf/iclr/Yang0LLLZC24} and MIPRO \citep{DBLP:conf/emnlp/Opsahl-OngRPBPZ24} sample $N$ prompt configuration candidates and score each by replaying the pipeline on a mini-batch of training set. However, re-running a memory system requires feeding the entire long interaction trajectory in an online manner, making the computational cost increasingly prohibitive as the number of prompt candidates grows.

\paragraph{Textual back-propagation.} Methods such as TextGrad \citep{DBLP:journals/nature/YuksekgonulBBLLHGZ25} avoid replaying the trajectory by propagating natural-language feedback backward through the computation graph. However, because the computation graph is extremely long and many operations are unrelated to the failure, it is highly susceptible to signal blockage, downstream over-correction, and upstream pollution \citep{Huang2026TextResNetDA}.

\paragraph{Common root cause.} The three failure modes share a single root cause: each family attempts to reason about, propagate signals through, or replay the \emph{entire} causal chain end-to-end. Our approach instead first localizes the faulty operation on the execution graph, reducing prompt optimization to a small, well-scoped sub-problem on which any of the three families above can be applied without modification.

\subsection{Experimental Details}
\label{app:opt-details}

\paragraph{Setup.} For Mem0, we optimize three prompts: the fact-extraction prompt used during memory construction, the memory-update decision prompt, and the question-answering prompt used at inference time. We assume a developer-in-the-loop setting in which the developer understands the target memory system and has access to data for iterative evaluation and improvement. The configuration for running Mem0 and collecting execution graphs follows Appendix~\ref{app:memory_system_setup}. Mem0 is initialized with its default prompt configuration. Both \memtrace and the prompt optimizer use GPT-5.4 as the backbone.

We randomly sample three LoCoMo user trajectories for optimization and reserve the remaining seven for testing. Optimization proceeds for three rounds. At round $j$, we run the current memory system on the $j$-th trajectory and evaluate it using the corresponding question-answer pairs.

\paragraph{\memtrace-Guided Optimization.}
Failed cases are passed to \memtrace for operation-level credit assignment. We initialize its to-explore list with the source evidence associated with each case and provide a high-level description of the Mem0 pipeline as prior knowledge. After \memtrace localizes the faulty operation, optimization is restricted to the prompts participating in that operation. 

Because each localized operation subgraph is small, we use a lightweight prompt optimizer. It first generates feedback from the failure information and operation subgraph, aggregates the feedback across cases, and then rewrites the target prompt. Following TextGrad, the optimizer retains one previous version of each prompt variable. After each round, we update the corresponding Mem0 prompts before proceeding to the next trajectory.

\paragraph{No-Attribution Baseline.}
To isolate the contribution of operation-level credit assignment, we construct a no-attribution baseline using the same optimization data, failure cases, backbone, optimizer, and three-round budget. This baseline bypasses \memtrace localization and provides the optimizer only with the final response-stage subgraph. Since no faulty operation is identified, the optimizer is asked to jointly update all three prompts. Therefore, the only difference between the two settings is whether localized credit assignment is available.

\paragraph{Optimization Cost.}
Table~\ref{tab:prompt_optimization_cost} reports the average cost of our closed-loop optimization pipeline. Overall, the cost is dominated by \memtrace. Nevertheless, its average wall-clock runtime is only 1.33 minutes per failed case, making it practical for an offline prompt optimization loop. The subsequent optimizer stages are lightweight because they operate on localized operation subgraphs and target only the prompts involved in the faulty operation.

% \begin{table}[t]
% \centering
% \small
% \begin{tabular}{lccc}
% \toprule
% Dataset & Sparse & Dense & Hybrid \\
% \midrule
% LoCoMo & 78.48 & 79.75 & \textbf{89.87} \\
% LongMemEval & 70.27 & 79.73 & \textbf{81.76} \\
% RealMem & 34.85 & 31.82 & \textbf{39.39} \\
% \midrule
% Overall & 61.20 & 63.77 & \textbf{70.34} \\
% \bottomrule
% \end{tabular}
% \caption{\textbf{The performance of different retrieval methods across datasets.} All values are reported as percentages. ``Overall'' aggregates results across all datasets.}
% \label{tab:retrieval_recall}
% \end{table}

% \begin{table}[t]
% \centering
% \small
% \begin{tabular}{lc}
% \toprule
% Retrieval Query & Recall@8 \\
% \midrule
% Query & 59.50 \\
% Query + System Prediction & 62.20 \\
% Query + Golden Answer & \textbf{70.34} \\
% \bottomrule
% \end{tabular}
% \caption{\textbf{Source-evidence retrieval performance under different query-construction strategies.} All values are reported as percentages.}
% \label{tab:retrieval_query_construction}
% \end{table}

\begin{table}[t]
\centering

\begin{subtable}[t]{0.48\columnwidth}
\centering
\small
\begin{tabular}{lccc}
\toprule
Dataset & Sparse & Dense & Hybrid \\
\midrule
LoCoMo & 78.48 & 79.75 & \textbf{89.87} \\
LongMemEval & 70.27 & 79.73 & \textbf{81.76} \\
RealMem & 34.85 & 31.82 & \textbf{39.39} \\
\midrule
Overall & 61.20 & 63.77 & \textbf{70.34} \\
\bottomrule
\end{tabular}
\caption{The performance of different retrieval methods across datasets.}
\label{tab:retrieval_recall}
\end{subtable}
\hfill
\begin{subtable}[t]{0.48\columnwidth}
\centering
\small
\begin{tabular}{lc}
\toprule
Retrieval Query & Recall@8 \\
\midrule
Query & 59.50 \\
Query + System Prediction & 62.20 \\
Query + Golden Answer & \textbf{70.34} \\
\bottomrule
\end{tabular}
\caption{Source-evidence retrieval performance under different query-construction strategies.}
\label{tab:retrieval_query_construction}
\end{subtable}

\caption{\textbf{Retrieval performance analysis.} All values are reported as percentages.}
\label{tab:retrieval_analysis}
\end{table}

\begin{figure*}[t!]
\centering
\begin{subfigure}[t]{0.44\textwidth}
  \centering
  \includegraphics[
    width=\linewidth,
    height=0.32\textheight,
    keepaspectratio
  ]{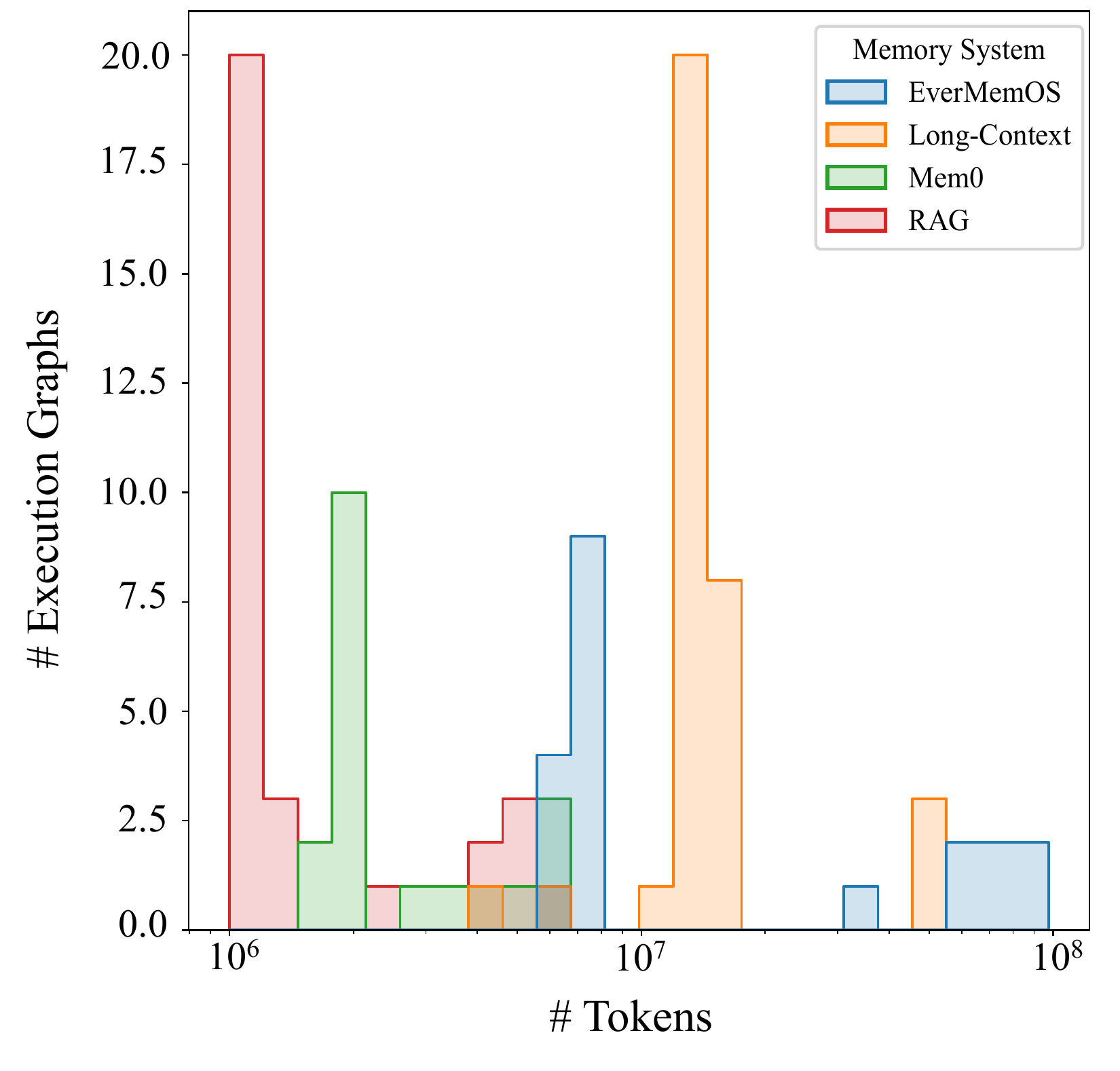}
  \caption{Token Distribution}
  \label{fig:graph_token_distribution_by_system}
\end{subfigure}\hfill
\begin{subfigure}[t]{0.52\textwidth}
  \centering
  \includegraphics[width=\linewidth]{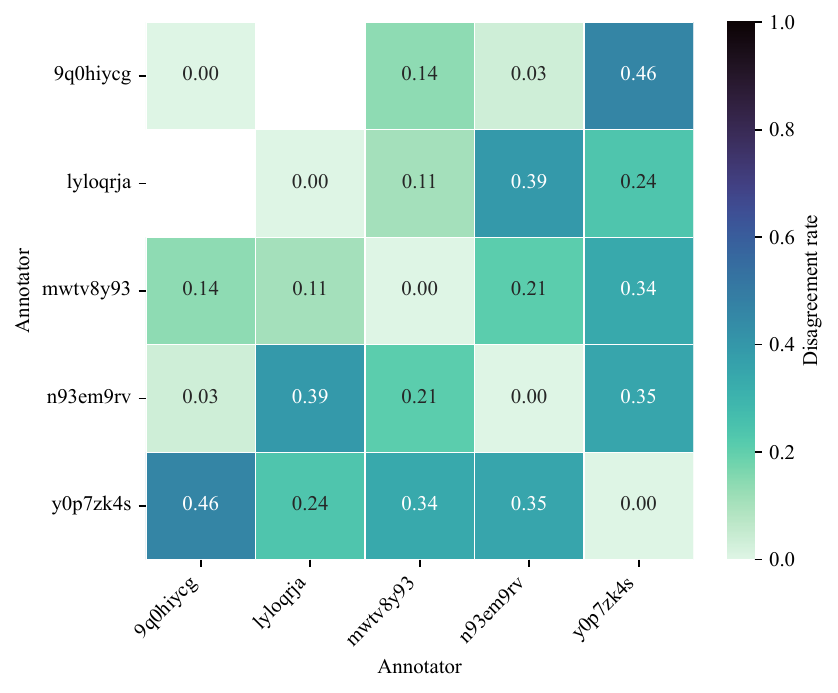}
  \caption{Annotator Disagreement}
  \label{fig:annotator_disagreement}
\end{subfigure}

\caption{{\bf Additional dataset analysis.}
(a) Token distribution of execution graph logs for each memory system.
(b) Pairwise disagreement rates among annotators. Darker colors indicate higher disagreement.
The disagreement between annotators 9q0hiycg and lyloqrja cannot be computed because their annotated cases do not overlap.}
\label{fig:additional_dataset_analysis}
\end{figure*}

\section{Additional Analysis}

% \subsection{Retrieval Performance}
% \label{app:retrieval_performance}

% We further examine whether concatenating the original question with the golden answer can effectively retrieve source evidence for initializing graph exploration. We compare three retrieval strategies including BM25-based sparse retrieval, dense retrieval with Qwen-Embedding-4B, and RRF-based hybrid retrieval used by \memtrace. We report retrieval performance using Recall@8. As shown in Table~\ref{tab:retrieval_recall}, hybrid retrieval performs best across all datasets. On LoCoMo and LongMemEval, even the weakest retrieval method exceeds 70\%, suggesting that golden answers provide strong cues for locating relevant source messages. For RealMem, dense retrieval underperforms sparse retrieval, possibly because messages from RealMem are longer, a setting where sparse retrieval can be more reliable~\citep{DBLP:journals/tacl/LuanETC21}. Overall, these results show that using golden answers yields high-quality starting points for \memtrace exploration.

\subsection{Retrieval Query Construction}
\label{app:retrieval_query_construction}

Golden answers provide strong cues for retrieving source evidence, but they are generally unavailable when diagnosing failures in production. We therefore evaluate two deployment-compatible alternatives: using the user query alone and concatenating the query with the system prediction. The latter is always available in the execution trace, even when the prediction is incorrect. We keep the RRF-based hybrid retriever fixed across all settings and report source-evidence Recall@8.

As shown in Table~\ref{tab:retrieval_query_construction}, the golden answer achieves the highest recall, as expected. Nevertheless, incorporating the system prediction improves Recall@8 from 59.50\% to 62.20\% over using the query alone. Although the prediction is incorrect, it may still contain relevant entities, or partial facts derived from the memory system, providing useful cues for locating the corresponding source messages. This effect varies across memory systems. Adding the system prediction improves retrieval on Mem0 and EverMemOS, but degrades it on RAG and Long-Context. For example, Recall@8 on Long-Context decreases from 70.92\% to 64.42\%. This suggests that, for advanced memory systems, model predictions provide a practical alternative when golden answers are unavailable. In production, we can further use user feedbacks to identify relevant source evidence, which we leave as an interesting direction for future work.

\subsection{Additional Dataset Analysis}
\label{app:dataset_analysis}

\paragraph{Memory system execution logs exceed the long-context window of current popular LLMs.} As presented in Figure~\ref{fig:graph_token_distribution_by_system}, each system produces traces exceeding one million tokens, with more advanced systems approaching $10^{7}$ tokens. Moreover, long-context memory further increases log size because each update operation records both the pre- and post-update context windows. 
As a result, the scale of these traces makes it difficult for practitioners to process directly with long-context LLMs for end-to-end inspection  (see Table~\ref{tab:attribution_accuracy} and Appendix~\ref{app:failure_attribution_details}).

\begin{wrapfigure}{r}{0.60\columnwidth}
    \centering
    \includegraphics[width=\linewidth]{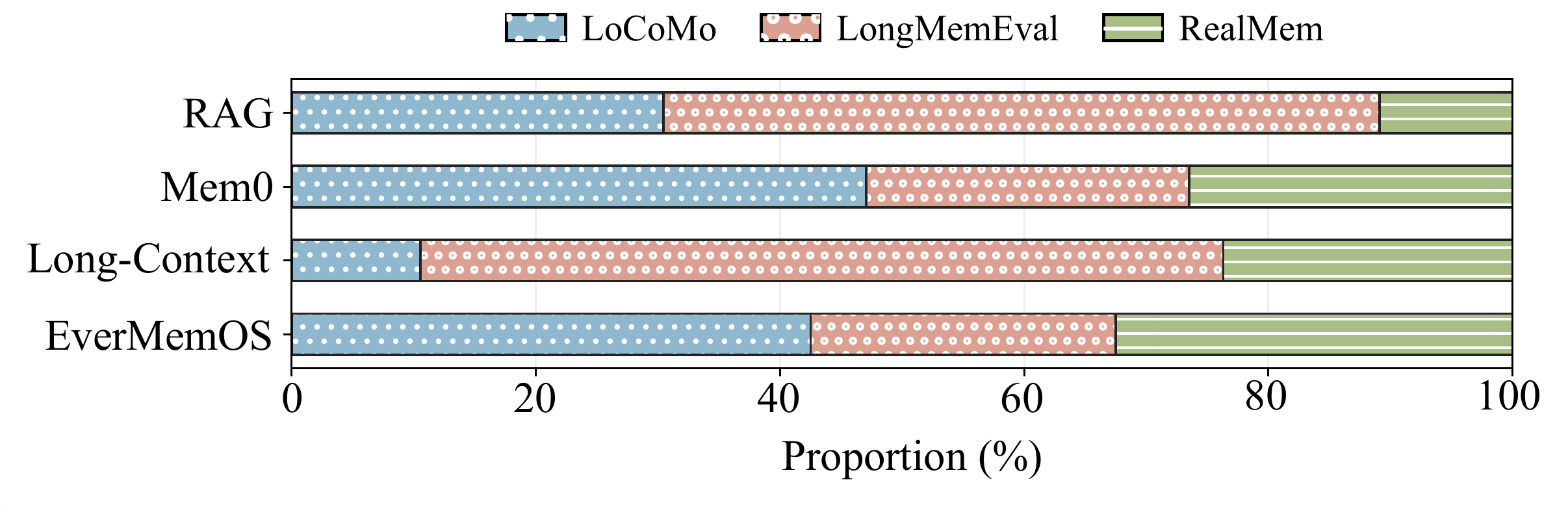}
    \caption{The dataset source distribution of system-related errors for each memory system.}
    \label{fig:system_error_dataset_distribution}
\end{wrapfigure}

\paragraph{Data annotation is intrinsically challenging.} We compute pairwise annotator disagreement based on the first-round annotations. As shown in Figure~\ref{fig:annotator_disagreement}, annotators exhibit non-trivial disagreement, with pairwise disagreement rates ranging from 3\% to 46\%. This suggests that annotating failure attribution cases is intrinsically challenging, especially for Mem0 and EverMemOS cases. 

\paragraph{System-related errors exhibit different dataset-source distributions across memory systems.} As shown in Figure~\ref{fig:system_error_dataset_distribution}, the errors exhibit two different distribution patterns across memory systems. 
For Mem0 and EverMemOS, LoCoMo contributes the largest share of system-related errors, whereas for RAG and long-context memory, most errors come from LongMemEval.
One possible reason is that LongMemEval contains long user trajectories for each question-answering case, introducing more distracting context and making direct retrieval or long-context reasoning more difficult.

\clearpage 

\begin{table*}[t]
\centering
\small
\begin{tabularx}{\textwidth}{@{}X@{}}
\toprule
\textbf{Systematic Error Analysis of Mem0} \\
\midrule
This analysis examines failure modes in Mem0 across memory construction, maintenance, retrieval, and downstream question answering, with the goal of understanding why conversationally available information does not reliably reappear as correct personalized recall. The failures show that robust performance depends not only on storing plausible memories, but on preserving the exact parts of an utterance that later become answer-bearing keys: temporal anchors, identifying nouns, updated numeric values, quoted formulations, evaluative wording, conversational next steps, completion states, speaker-specific commitments, and source details such as where an item was obtained. Across the cases, four recurring weaknesses stand out. Memory updates can corrupt previously correct facts by rewriting temporal grounding, collapsing repeated but distinct events into a single record, or stripping away relational and provenance details needed for later comparisons or literal recall. Fact extraction can prevent critical information from ever entering memory, especially when the decisive content is a relative date, a timing clause, a negative preference, a short evaluative phrase, a specific emotion, a structured plan, or an assistant-provided list that later becomes the target of recall. Retrieval based on the raw user question often favors semantically adjacent or person-related memories while omitting the exact memory whose wording, entity label, recency, completeness, or set membership resolves the query. Finally, even when relevant evidence is present, the answering model does not consistently use it to perform literal recall, enumerate all qualifying items, or preserve necessary temporal and state distinctions, and instead drifts toward plausible but unsupported continuations or incomplete counts. 

\medskip
\hspace{2em}A first category of failures arises from memory maintenance errors, particularly update operations that preserve surface topicality while silently overwriting essential event structure or fact specificity. In one pattern, \redbold{a memory can remain textually similar yet have its temporal grounding reassigned to a later timestamp}, as when an artwork memory about something created ``last week'' was rebound from the correct August context to October. In another pattern, \redbold{repeated events with similar schemas are wrongly merged instead of stored as separate episodes}. Dave’s car-show history illustrates this clearly: an earlier memory that he went to a car show ``last weekend'' was updated into ``last Friday'' after a later message about a different car-show visit, effectively replacing one attendance event with another. A related subtype is specificity-degrading update. \redbold{Here the system starts from a locally correct memory and then rewrites it into a broader, less answerable version}, as when ``User has opened their own car maintenance shop'' became merely ``Opened a shop which was a dream and involved a lot of hard work''. The newer travel case showed that this degradation can remove relational detail needed for comparative reasoning, and the tennis-racket case shows an adjacent provenance-loss variant in which an update deletes the exact source field that later answers a ``where from'' question. ``Has a new tennis racket from a sports store downtow'' was rewritten into ``Currently focusing on improving tennis game with a new racket'', preserving topic but erasing the purchase location. This is not just stylistic drift. Once update decisions treat memories as mutable summaries rather than protected records with answer-bearing fields, the store loses timestamp integrity, event cardinality, lexical specificity, relational comparability, and source provenance, making later recall systematically unreliable.

\\
\bottomrule
\end{tabularx}
\caption{\textbf{Full systematic error analysis report automatically generated for Mem0.} Red bold text highlights major recurring failure mechanisms. Since \memtrace is not perfectly accurate, some identified errors may contain minor inaccuracies. However, from an overall perspective, \memtrace still provides meaningful insights.}
\label{tab:mem0-error-analysis}
\end{table*}

\begin{table*}[t]
\centering
\small
\begin{tabularx}{\textwidth}{@{}X@{}}
\toprule
\textbf{Systematic Error Analysis of Mem0} \\
\midrule

\hspace{2em}A second major category consists of extraction failures, where decisive information never becomes retrievable because it is not converted into memory units at all, or is converted only in a weakened form. \redbold{One subtype is policy-induced omission: the extractor is explicitly instructed to store only user-message facts, so assistant-provided content such as a previously given example answer, a detailed plan, an itinerary revision, a recipe ingredient list, a delivered module breakdown, a venue recommendation list, or the declared next component after a milestone is dropped even though later questions ask the system to remember exactly what the assistant had said or already provided.} The Portland music-venue failure makes this especially clear: the assistant’s full ordered list ending with ``Revolution Hall'' was fully present in the conversation but extraction returned an empty list, so later recall became impossible. A particularly important form of this mismatch is assistant-originated state/progress omission, where the omitted content is not merely descriptive but records that something has already been pushed, locked in, delivered, or advanced to the next stage; once that state transition is absent from memory, later responses regress to outdated planning talk or redundant re-explanation. \redbold{A second subtype is subtle user-fact omission, where the extractor returns no facts even though the user states a durable preference, constraint, or evaluative takeaway that is crucial for future responses.} \redbold{A third subtype is temporal-clause omission: when a message contains both a descriptive fact and a date-bearing or relative-time phrase, extraction tends to retain the high-level description while dropping the temporal detail that later becomes the direct target of a ``when'' question.} \redbold{A fourth subtype is evaluative-detail omission in socially directed messages, where praise, characterization, or causal attributions are compressed into generic encouragement or discarded altogether.} \redbold{A fifth and increasingly important subtype is structured-payload omission. In these cases the source message contains a long but highly answerable payload—such as a workout protocol, a route update, direct manifestations of a narrative pressure point, a paper breakdown, an ingredient list with exact quantities, or an ordered venue list—yet extraction returns an empty list because the content is not framed as a simple user profile fact.} A broader extraction pathology is answer-bearing lexical compression. \redbold{Here the extractor preserves topical gist but drops the exact word, phrase, or structured item that the later question asks for.} Dave’s statement that fixing cars “makes me proud” was reduced to feeling ``great'', and his description of working on cars as ``like therapy'' was rewritten into a more generic engine-focused formulation. These are not harmless paraphrases. Because Mem0 later searches with raw user questions, losing exact emotion terms, business labels, milestone transitions, module-delivery states, numeric quantities, schedule revisions, ordered list items, or procedural details weakens both retrieval match quality and answer precision.

\medskip
\hspace{2em}Retrieval errors remain a major source of failure and reveal several related weaknesses in using the raw question as a direct search query over fragmented memory units. \redbold{One recurring pattern is companion-memory omission, where the system retrieves a memory describing the focal event but not the adjacent fact needed to answer the question fully.} \redbold{Another is semantic genericity failure, in which broad memories about the same person, topic, or activity outrank the decisive, more specific memory.} Questions about counseling services, maintenance-program follow-up, networking plans, detailed workout instructions, KudiFlow route updates, and clothing pickup obligations all show that topical overlap alone is insufficient: generic memories about support, planning, flexibility, wardrobe organization, route changes, or KudiFlow tips can dominate while the single answer-bearing next step, named entity, actual plan payload, or locked-in revision is absent. \redbold{A third subtype is identifying-label omission under dense topical competition.} In the Ferrari case, the store already contained the decisive memory that Calvin had recently gotten a Ferrari and called it a ``masterpiece on wheels'', yet retrieval returned only vague vehicle-related, motivational, and person-specific memories. A fourth subtype is aggregate-set omission. \redbold{When the question asks for a list or inventory of activities or obligations, retrieval does not reliably surface the multiple distinct episodic memories needed to compose the set.} In the friend-activities case, the store contained separate memories for park walks, a countryside road trip, a shop-opening celebration, and card-playing nights, yet the retrieved context consisted of off-target items such as catch-up plans, generic positive interactions, and self-expression through cars. The clothing-store case shows the same failure in obligation form: the system retrieved the blazer dry-cleaning pickup but omitted both Zara boots memories, so a question requiring the total number of items to pick up or return was collapsed to a single item. \redbold{A fifth subtype is stale-memory preference despite available updates.} In the bird-species case, retrieval surfaced the older count of 27 and a related Northern Flicker memory, but omitted the later memory explicitly storing the updated total of 32. A sixth subtype, highlighted by the guide-editing failure, is status-contrast omission. Some questions do not merely ask for a topic but for the current state of multiple requested edits, such as whether one component has already been added while another has not. In those cases, \redbold{retrieval surfaces positive-progress memories around the project but misses the decisive contrasting evidence that distinguishes completed from not-yet-completed items}. Across these cases, direct vector search appears biased toward person identity, broad topical similarity, and diffuse semantic neighborhoods, while underweighting answer-bearing nouns, updated totals, quoted aliases, enumerative coverage, revision state, milestone transitions, event-to-consequence links, and the need to retrieve complete sets of related obligations or events rather than just the most individually similar item.

\\
\bottomrule
\end{tabularx}
\caption{The continued part of Table~\ref{tab:mem0-error-analysis}.}
\label{tab:mem0-error-analysis-continued-1}
\end{table*}

\begin{table*}[t]
\centering
\small
\begin{tabularx}{\textwidth}{@{}X@{}}
\toprule
\textbf{Systematic Error Analysis of Mem0} \\
\midrule

\hspace{2em}Question-answering errors show that evidence inclusion alone is not sufficient unless the answering model reliably identifies, combines, and computes over the memories that directly resolve the question. \redbold{One subtype is evidential selection failure, where the model prefers a more elaborate but less relevant narrative over a concise decisive memory, producing answers that are thematically plausible yet unsupported by the best evidence in context.} \redbold{Another subtype is task-mapping failure: even when the retrieved context contains the right project state and next-step cues, the model may answer a neighboring but different question by drifting toward another salient thread.} The script-development case is illustrative: the context already contained ``optimizing engagement through pacing'' and ``ensure efficient data throughput in the story'', yet the answer shifted to an unrelated Act III resolution task rather than using the retrieved evidence to name pacing as the next priority. \redbold{A third subtype is temporal grounding failure: when context contains relative-time expressions paired with message timestamps, the model may answer with the message date itself or anchor to a different but topically related memory rather than converting the relative expression into the correct calendar interval.} The art-events case highlights \redbold{a further subtype that is better characterized as aggregation and counting failure}. There, the final context already contained multiple qualifying art-event memories within the target window, but the model enumerated only one event and concluded the answer was 1 instead of 4. This shows that even when retrieval succeeds, the QA stage may fail to scan the full context, apply the query's inclusion criteria consistently, or aggregate all matching memories into a correct total. The failures also continue to show relative-time arithmetic and update-tracking weakness even when the needed evidence is partly present, with the model tending to copy the most salient duration string or stale numeric value rather than perform the small subtraction, filtering, or state resolution required by the question. More broadly, when the exact memory is absent or degraded, the QA model falls back to semantic reconstruction. In the maintenance-program case it improvised a plausible checklist rather than recalling the established next step; in the Ferrari case it inferred only that Calvin got some unspecified vehicle; in the emotion case it reproduced the extractor’s weakened wording and answered ``great'' instead of ``proud''; and in the workout-plan case it generated a generic seven-day routine from high-level fitness constraints rather than reproducing the stored three-day protocol. The guide-editing failure reveals a sharper completion bias: when retrieved context contains evidence that progress has been made on a project but omits the explicit memory that some requested detail is still missing, the model tends to over-affirm full completion rather than preserve partial status. These errors are often downstream manifestations of earlier pipeline failures, but they also reveal an independent weakness: the model does not consistently distinguish literal recall from plausible thematic synthesis, nor does it reliably perform complete set aggregation or abstain when the retrieved evidence lacks the exact answer.

\medskip
\hspace{2em}Overall, the failures point to a pipeline whose reliability depends on aligning memory construction policy with the kinds of conversational recall it is expected to support, preserving episode identity and timestamp integrity during updates, protecting answer-bearing specificity during paraphrase, extracting structured and assistant-originated content rather than only high-level user summaries, retrieving the minimal set of decisive memories rather than semantically adjacent background facts, and constraining generation to explicit evidence and simple verifiable inference. The most important implication is that errors are often compositional: a restrictive extraction policy can make later recall impossible, a paraphrastic extraction can preserve topic while erasing the crucial word, a null extraction on a long structured or enumerated message can prevent any of the answer-bearing payload from entering memory, an update can silently collapse two distinct experiences into one or strip away the relational, lexical, or provenance detail that supports comparison and literal recall, a retrieval layer optimized for topical similarity can still miss the memory that explicitly names the answer, contains the latest value, or completes the relevant set, and a QA model can still ignore or undercount the correct evidence even when it is present. Effective personalized QA therefore requires better handling of assistant-originated conversational content when relevant, stronger preservation of temporal expressions, numeric updates, negative knowledge, evaluative wording, exact emotion terms, ingredient quantities, itinerary revisions, delivery states, venue lists, purchase sources, and next-step plans, update rules that protect event boundaries as well as lexical, relational, and provenance specificity, retrieval mechanisms that reward answer-bearing recency, completeness, set coverage, and state contrast over generic semantic proximity, and more conservative answer synthesis that explicitly acknowledges when the stored evidence does not support exact recall or when light filtering, counting, or aggregation is required.

\\
\bottomrule
\end{tabularx}
\caption{The continued part of Table~\ref{tab:mem0-error-analysis}.}
\label{tab:mem0-error-analysis-continued-2}
\end{table*}

\begin{table*}[t]
\centering
\small
\begin{tabularx}{\textwidth}{@{}X@{}}
\toprule
\textbf{Systematic Error Analysis of EverMemOS} \\
\midrule
This analysis examines failure modes in EverMemOS across memory construction, retrieval control, sufficiency assessment, and downstream answer generation, with the aim of identifying the system behaviors that most often break end-to-end factual recall. The dominant errors do not arise simply because information was never stored. Instead, many failures reflect losses of precision at stage boundaries: the system may preserve the decisive memory in storage, yet retrieve the wrong stage of the user’s plan, judge incomplete evidence as sufficient, or pass a misleading context to the answer model. Across the observed cases, the most consequential weakness is therefore not raw memory coverage but state fidelity. EverMemOS frequently retains enough topical material to sound plausible, while missing the exact fact, relation, temporal anchor, commitment status, or latest event update needed to answer correctly.

\medskip
\hspace{2em}A major class of errors comes from response-stage evidence misuse, where the retrieved context already contains the key information but the question-answering model still produces the wrong answer. These failures appear in several recurring forms. \redbold{One is competitive salience failure, where the model latches onto a nearby but non-answer-bearing topic and substitutes it for the requested target.} \redbold{Another is schema mismatch, where the model answers under the wrong counting or comparison frame, such as converting a question about quantities of items into a count of actions.} \redbold{A third is attribute misbinding, where the right cluster of candidate facts is present but a distinctive attribute, example, or description is attached to the wrong entity.} \redbold{A fourth is premature commitment on unresolved states: when memory preserves an open choice or a not-yet-completed action, the answer model collapses that uncertainty into a falsely finalized plan.} These patterns show that EverMemOS can often deliver enough evidence to support the answer in principle, yet the final generation stage remains brittle when success depends on preserving exact semantic roles and decision states rather than producing a topically plausible response.

\medskip
\hspace{2em}A second major category involves retrieval steering and evidence-selection failures inside the adaptive retrieval pipeline. In some cases, relevant memories are stored correctly and even appear in the broader hybrid candidate pool, but the system’s controller elevates the wrong evidence into the sufficiency-check set or final QA context. \redbold{One subtype is stale-state exclusion: the decisive memory is the latest update to a schedule, booking status, or plan state, but reranking favors older, topically similar memories, causing the system to answer from superseded evidence.} This is especially damaging for questions about what is next, what is already booked, or whether a plan has been confirmed, because older preparatory memories remain semantically close while differing critically in status. \redbold{A second subtype is off-target Round-2 query generation.} When Round-1 evidence is judged insufficient, EverMemOS can misdiagnose what is missing and generate follow-up queries that overfit to salient but irrelevant fragments of the current pool. In the clearest cases, the system imports the wrong retrieval schema entirely—for example, treating a non-temporal planning query as though it were a temporal-interval reconstruction problem—and then expands follow-up queries around downloading, scheduling, or setup details instead of the actual revision-decision episode. These failures reveal that retrieval quality is constrained not only by candidate recall, but by how the system interprets insufficiency and decides which latent subproblem to search for next.

\\
\bottomrule
\end{tabularx}
\caption{\textbf{Full systematic error analysis report automatically generated for EverMemOS.} Red bold text highlights major recurring failure mechanisms. Since \memtrace is not perfectly accurate, some identified errors may contain minor inaccuracies. However, from an overall perspective, \memtrace still provides meaningful insights.}
\label{tab:evermemos-error-analysis}
\end{table*}

\begin{table*}[t]
\centering
\small
\begin{tabularx}{\textwidth}{@{}X@{}}
\toprule
\textbf{Systematic Error Analysis of EverMemOS} \\
\midrule

\hspace{2em}A third recurring category is temporal and state-tracking fragility, which cuts across retrieval, sufficiency checking, and answer generation. Many questions depend not merely on topical relevance but on recovering the correct event state at the correct point in time: distinguishing current schedule from prior schedule, confirmed booking from intended booking, or today’s planned workout from a remembered but misaligned future reference. EverMemOS struggles with several variants of this problem. \redbold{It can confuse mention time with event time, fail to privilege the most recent update over earlier planning memories, or collapse distinctions between intention, recommendation, and completion.} In other cases, \redbold{the sufficiency check itself is too coarse: it marks evidence as sufficient because the topic domain is covered, while overlooking a missing temporal or status-bearing fact that is essential to the question.} This leads the system to terminate retrieval early and answer from an incomplete context, as when intent to book campsites is treated as equivalent to campsites already being secured. These failures suggest that EverMemOS represents topical continuity more robustly than event-state transitions. It often remembers what the conversation was about, but not whether a key action was proposed, scheduled, postponed, completed, or still pending.

\medskip
\hspace{2em}Overall, the error patterns indicate that EverMemOS’s central challenge is not simply retrieving memories, but preserving the correct event state through retrieval control and answer grounding. The system is most reliable when a question can be answered by quoting a single salient fact, and least reliable when success depends on selecting the latest state update, distinguishing intention from completion, keeping unresolved decisions open, or generating follow-up retrieval plans that target the actual missing information rather than a plausible but irrelevant subtheme. Improving end-to-end performance will therefore require tighter coupling between retrieval, sufficiency judgment, and answer generation. Retrieval should optimize for evidential completeness at the level of event states and state transitions, not just topical relevance; the sufficiency checker should be more explicitly constrained to detect missing status and recency distinctions; and the final answer stage should be forced to ground claims in the retrieved temporal and decision structure. Without such improvements, the system will continue to fail even when the decisive memory is already present somewhere in storage or the initial candidate pool.

\\
\bottomrule
\end{tabularx}
\caption{The continued part of Table~\ref{tab:evermemos-error-analysis}.}
\label{tab:evermemos-error-analysis-continued-1}
\end{table*}

\begin{figure*}[t!]
\centering
\vfill 
\resizebox{1.0\textwidth}{!}{
\includegraphics{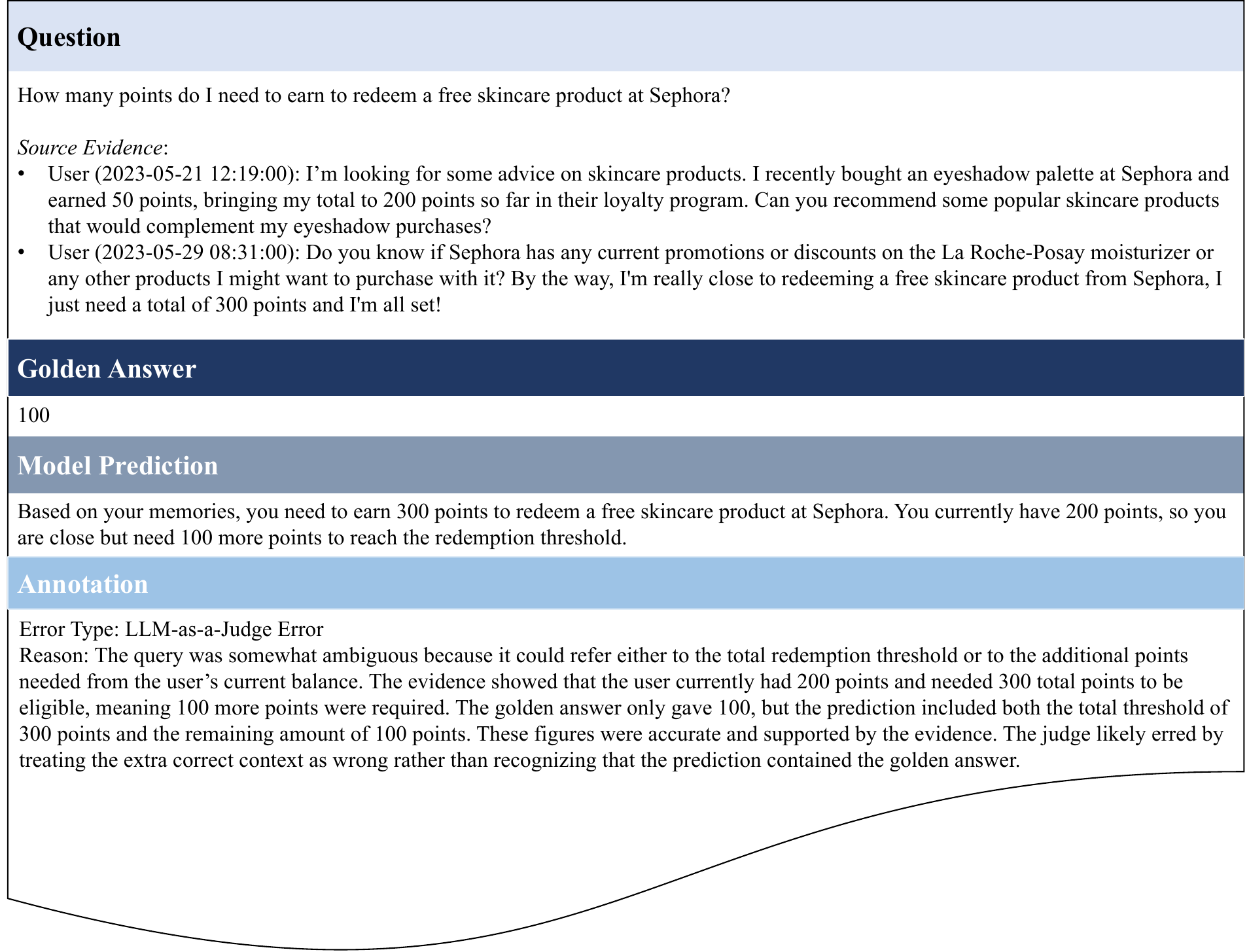}}
\caption{
\textbf{An LLM-as-a-Judge error case.}  Due to ambiguity between the total required points and the remaining points needed, the prediction provides both correct values (``300 total'' and ``100 remaining''), but the LLM judge marks it incorrect.
}
\vfill 
\label{fig:judge_error_case}
\end{figure*}

\begin{figure*}[t!]
\centering
\vfill 
\resizebox{1.0\textwidth}{!}{
\includegraphics{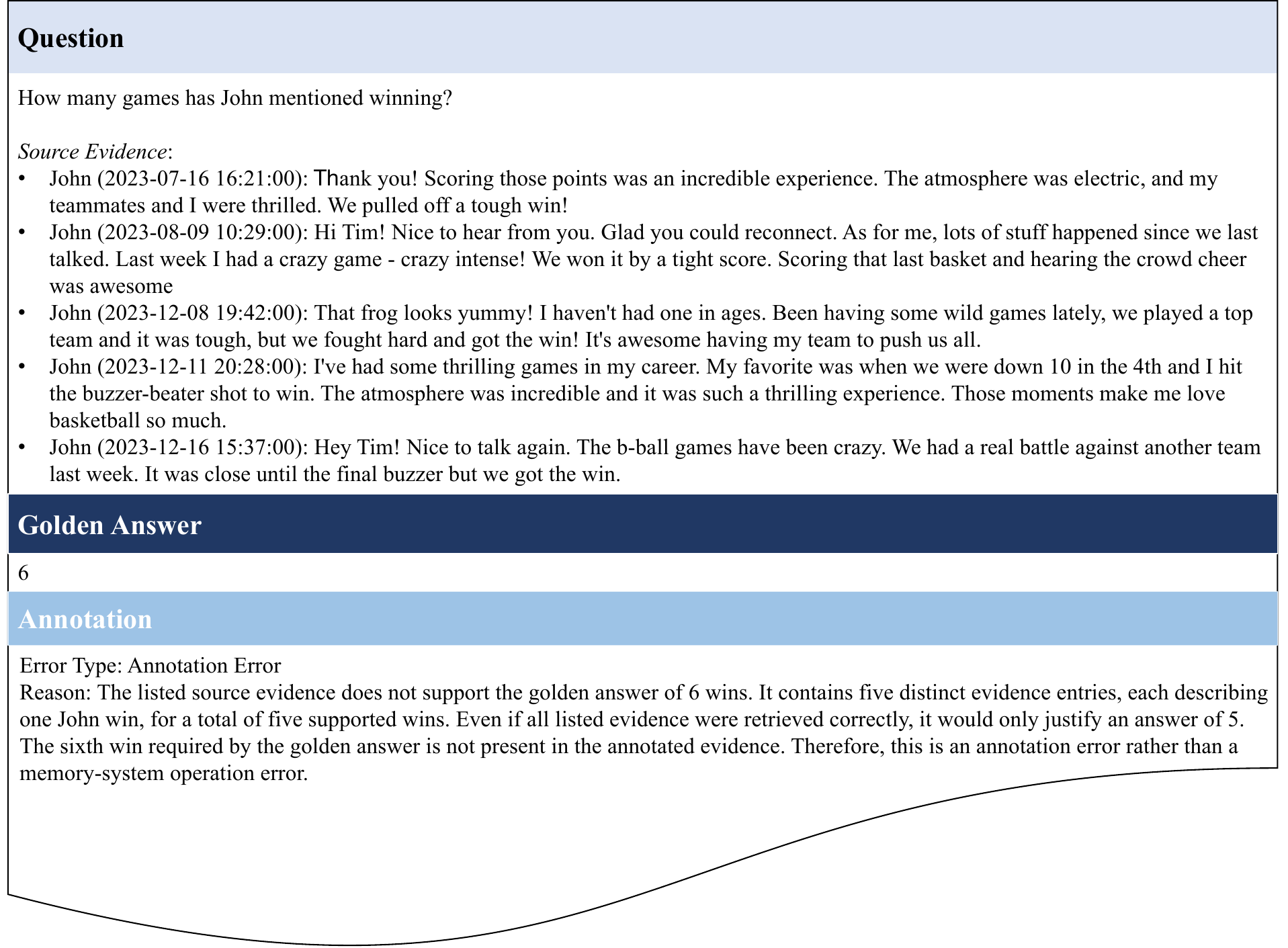}}
\caption{
\textbf{An Annotation case from LoCoMo.}  The evidence contains only five supported wins, but the golden answer expects six.
}
\vfill 
\label{fig:annotation_error_case_1}
\end{figure*}

\begin{figure*}[t!]
\centering
\vfill 
\resizebox{1.0\textwidth}{!}{
\includegraphics{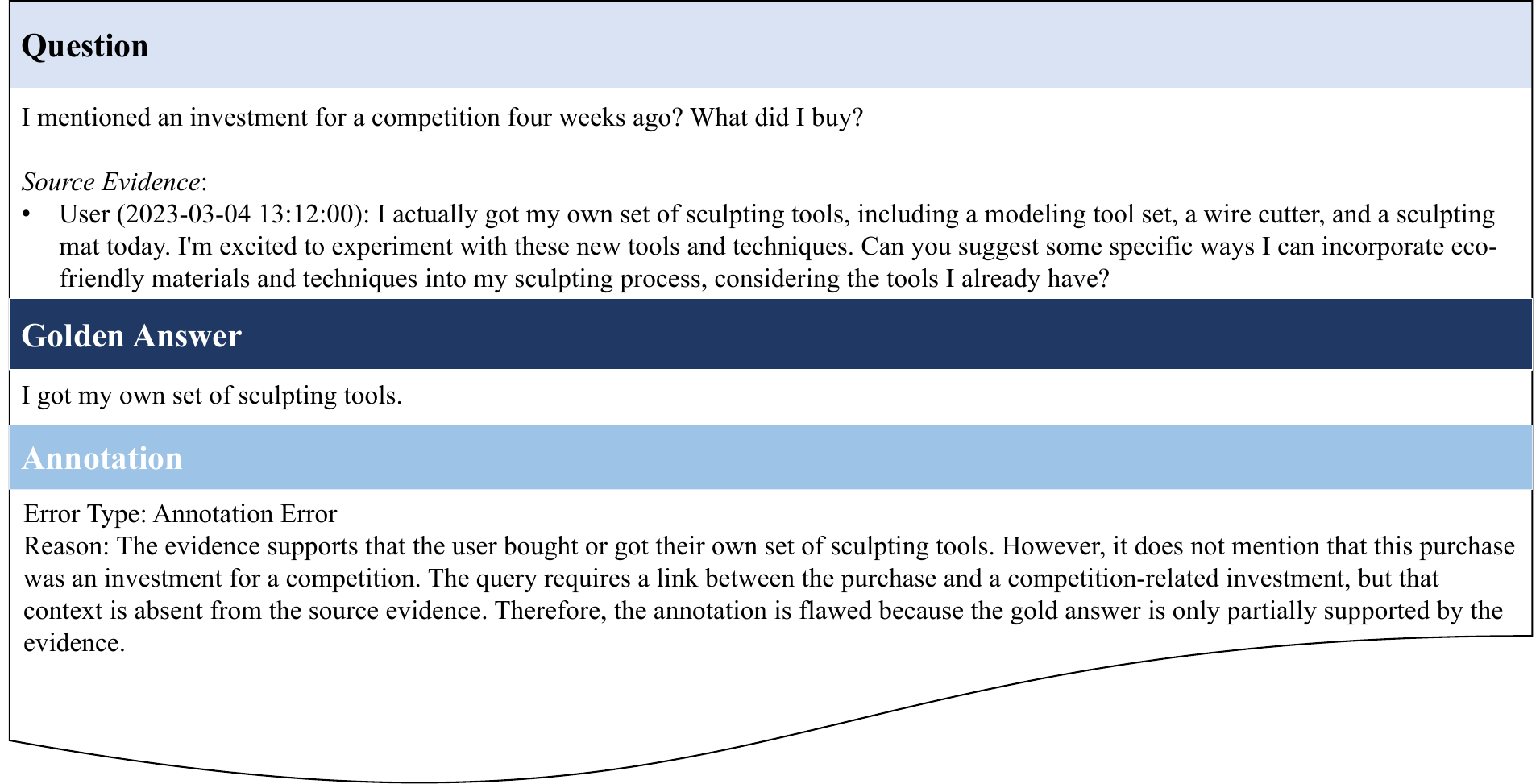}}
\caption{
\textbf{An Annotation case from LongMemEval.} The evidence confirms that the user obtains sculpting tools, but does not support the claim that the purchase is a competition-related investment. The question should instead remove the investment-related wording and ask a simpler supported query such as ``What did I buy four weeks ago''.
}
\vfill 
\label{fig:annotation_error_case_2}
\end{figure*}

\clearpage 

\clearpage

\begin{figure*}[p]
\centering
\hyperlink{fig:annotation-interface-overview-target}{\fbox{\textbf{Back to Overview}}}

\vspace{0.5em}
\includegraphics[width=\linewidth]{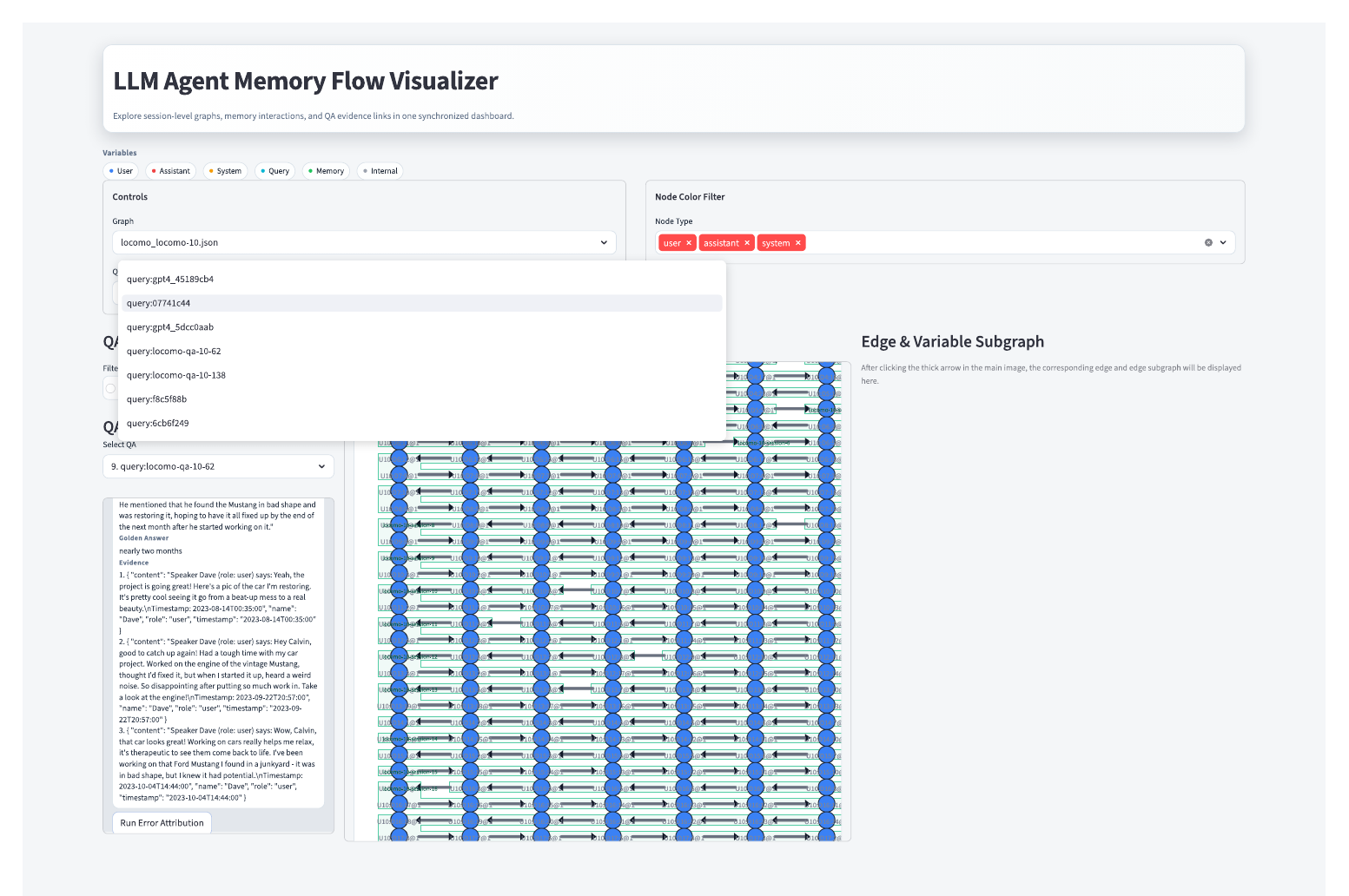}

\caption{\textbf{Full-size visualization of the annotation interface entry point.} This figure corresponds to the left thumbnail in the overview shown in Figure~\ref{fig:annotation-interface-overview}.}
\label{fig:annotation-interface-left}
\end{figure*}

\begin{figure*}[p]
\centering
\hyperlink{fig:annotation-interface-overview-target}{\fbox{\textbf{Back to Overview}}}

\vspace{0.5em}
\includegraphics[width=\linewidth]{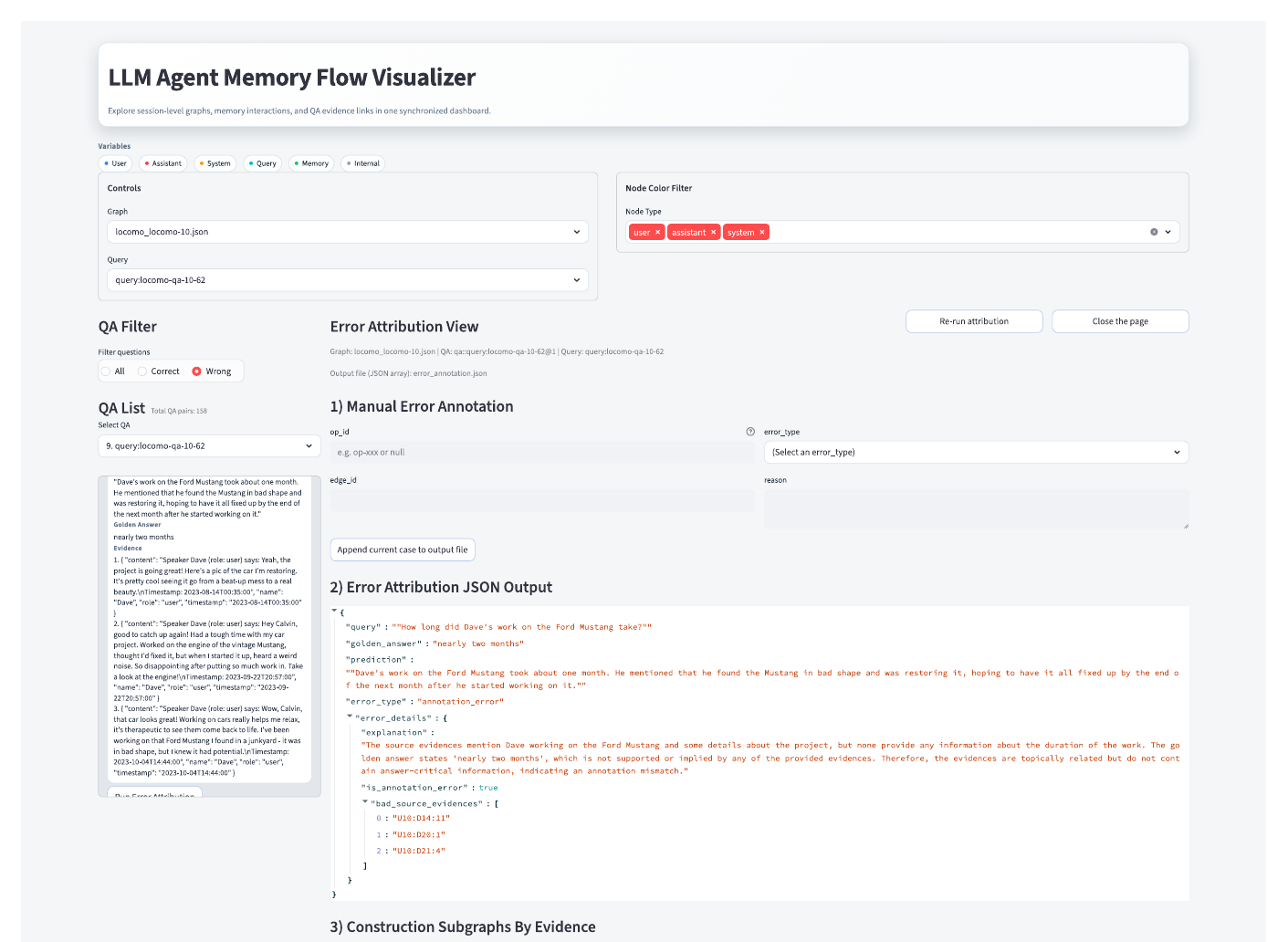}

\caption{\textbf{Full-size visualization of the annotation submission view.} This figure corresponds to the middle thumbnail in the overview shown in Figure~\ref{fig:annotation-interface-overview}.}
\label{fig:annotation-interface-middle}
\end{figure*}

\begin{figure*}[p]
\centering
\hyperlink{fig:annotation-interface-overview-target}{\fbox{\textbf{Back to Overview}}}

\vspace{0.5em}
\includegraphics[width=\linewidth]{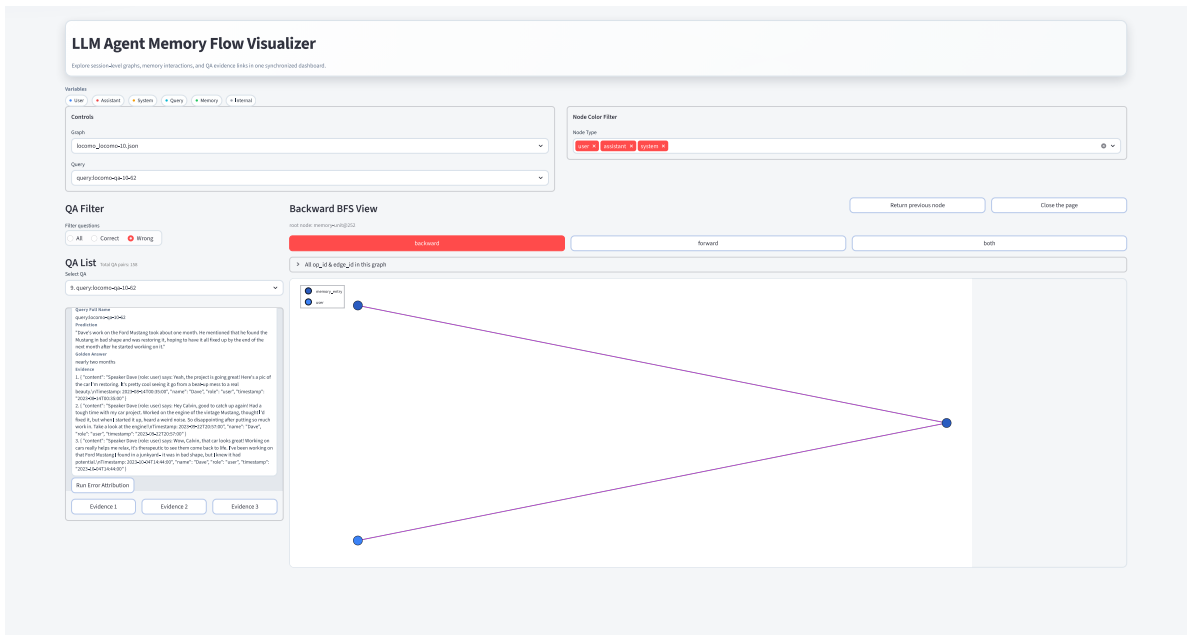}

\caption{\textbf{Full-size visualization of the execution graph exploration interface.} This figure corresponds to the right thumbnail in the overview shown in Figure~\ref{fig:annotation-interface-overview}.}
\label{fig:annotation-interface-right}
\end{figure*}